\documentclass{article}

\usepackage{microtype}
\usepackage{graphicx}
\usepackage{subfigure}
\usepackage{booktabs}
\usepackage{amsmath,amsthm}
\usepackage{amssymb}
\usepackage{bm}
\usepackage{braket,enumitem}

\newtheorem{proposition}{Proposition}

\usepackage{hyperref}

\usepackage[accepted]{icml2020}

\icmltitlerunning{Learning Curves in Kernel Regression and Wide Neural Networks}

\begin{document}

\twocolumn[
\icmltitle{Spectrum Dependent Learning Curves in Kernel Regression and Wide Neural Networks}



\icmlsetsymbol{equal}{*}

\icmlsetsymbol{equal}{*}

\begin{icmlauthorlist}
\icmlauthor{Blake Bordelon}{to}
\icmlauthor{Abdulkadir Canatar}{goo}
\icmlauthor{Cengiz Pehlevan}{to,cbs}
\end{icmlauthorlist}

\icmlaffiliation{to}{John A. Paulson School of Engineering and Applied Sciences, Harvard University, Cambridge, MA, USA}
\icmlaffiliation{goo}{Department of Physics, Harvard University, Cambridge, MA, USA}
\icmlaffiliation{cbs}{Center for Brain Science, Harvard University, Cambridge, MA, USA}


\icmlcorrespondingauthor{Cengiz Pehlevan}{cpehlevan@seas.harvard.edu}


\vskip 0.3in
]



\printAffiliationsAndNotice{}  

{\let\thefootnote\relax\footnote{{Code: \url{ https://github.com/Pehlevan-Group/NTK_Learning_Curves }}}}

\begin{abstract}
We derive analytical expressions for the  generalization performance of kernel regression as a function of the number of training samples using theoretical methods from Gaussian processes and statistical physics. Our expressions apply to  wide neural networks due to an equivalence between training them and kernel regression with the Neural Tangent Kernel (NTK). By computing the decomposition of the total generalization error due to different spectral components of the kernel, we identify a new spectral principle: as the size of the training set grows, kernel machines and neural networks fit successively higher spectral modes of the target function. When data are sampled from a uniform distribution on a high-dimensional hypersphere, dot product kernels, including NTK, exhibit learning stages where different frequency modes of the target function are learned. We verify our theory with simulations on synthetic data and MNIST dataset.
\end{abstract}

\section{Introduction}

Finding statistical patterns in data that generalize beyond a training set is a main goal of machine learning. Generalization performance depends on factors such as the number of training examples, the complexity of the learning task, and the nature of the learning machine. Identifying precisely how these factors impact the performance poses a theoretical challenge. Here, we present a theory of generalization in  kernel machines \cite{scholkopf_smola} and neural networks \cite{lecun2015deep} with wide hidden layers that addresses these questions.

The goal of our theory is not to provide worst case bounds on generalization performance in the sense of statistical learning theory \cite{vapnik1999overview}, but to provide analytical expressions that explain the average or a typical performance in the spirit of statistical physics. The techniques we use are a continuous approximation to learning curves previously used in Gaussian processes  \cite{sollich1999learning,sollich2001mismatch,sollich2002approx} and the replica method of statistical physics \cite{sherrington_spinglass,mezard1987spin}. 

We first develop an approximate theory of generalization in kernel regression that is applicable to any kernel. We then use our theory to gain insight into neural networks by using a correspondence between kernel regression and neural network training. When the hidden layers of a neural network are taken to infinite width with a certain initialization scheme, recent influential work \cite{jacot2018neural,arora2019exact,lee2019wide} showed that training a feedforward neural network with gradient descent to zero training loss is equivalent to kernel interpolation (or ridgeless kernel regression) with a kernel called the Neural Tangent Kernel (NTK) \cite{jacot2018neural}. 
Our kernel regression theory contains kernel interpolation as a special limit (ridge parameter going to zero).

Our contributions and results are summarized below:
\begin{itemize}[leftmargin=*,noitemsep,topsep=0pt]
    \item Using a continuous approximation to learning curves adapted from Gaussian process literature \cite{sollich1999learning,sollich2001mismatch}, we derive analytical expressions for learning curves for each spectral component of a target function learned through kernel regression. 
    \item We present another way to arrive at the same analytical expressions using the replica method of statistical physics and a saddle-point approximation \cite{sherrington_spinglass,mezard1987spin}. 
    \item Analysis of our theoretical expressions show that different spectral modes of a target function are learned with different rates. Modes corresponding to higher kernel eigenvalues are learned faster, in the sense that a marginal training data point causes a greater percent reduction in generalization error for higher eigenvalue modes than for lower eigenvalue modes. 
    \item When data is sampled from a uniform distribution on a hypersphere, dot product kernels, which include NTK, admit a degenerate Mercer decomposition in spherical harmonics, $Y_{km}$. In this case, our theory predicts that generalization error of lower frequency modes of the target function decrease more quickly than higher frequency modes as the dataset size grows. Different learning stages are visible in the sense described below.
    \item As the dimensions of data, $d$, go to infinity, learning curves exhibit different learning stages. For a training set of size $\ p \sim \mathcal{O}( d^l)$, modes with $k < l$ are perfectly learned, $k=l$ are being learned, and $k>l$ are not learned.
    \item We verify the predictions of our theory using numerical simulations for kernel regression and kernel interpolation with NTK, and wide and deep neural networks trained with gradient descent. Our theory fits experiments remarkably well on synthetic datasets and MNIST.
\end{itemize}

\subsection{Related Work}
Our main approximation technique comes from the literature on Gaussian processes, which is related to kernel regression in a certain limit. Total learning curves for Gaussian processes, but not their spectral decomposition as we do here, have been studied in a limited teacher-student setting where both student and teacher were described by the same Gaussian process and the same noise in  \cite{OpperVivarelli,sollich1999learning}.    
We allow 
arbitrary teacher distributions. Sollich also considered mismatched models where teacher and student kernels had different eigenspectra and different noise levels \cite{sollich2001mismatch}. The total learning curve from this model is consistent with our results when the teacher noise is sent to zero, but we also consider, provide expressions for, and analyze generalization in spectral modes.  We use an analogue of the ``lower-continuous" approximation scheme introduced in \cite{sollich2002approx}, the results of which we reproduce through the replica method \cite{mezard1987spin}. 

Generalization bounds for kernel ridge regression have been obtained in many contexts \cite{scholkopf_smola, cucker, vapnik1999overview, gyorfi}, but the rates of convergence often crucially depend on the explicit ridge parameter $\lambda$ and do not provide guarantees in the ridgeless case. Using a teacher-student setting, \citet{spigler2019asymptotic} showed that learning curves for kernel regression asymptotically decay with a power law determined by the decay rate of the teacher and the student. Such power law decays have been observed empirically on standard datasets \cite{hestness2017deep,spigler2019asymptotic}. Recently, interest in explaining the phenomenon of interpolation has led to the study of generalization bounds on ridgeless regression \cite{belkin2018understand, belkin_mitra_overfitting_perfect, belkin_interpolation_optimality, liang2018just}. Here, our aim is to capture the average case performance of kernel regression, as opposed to a bound on it, that remains valid for the ridgeless case and finite sample sizes. 

In statistical physics domain, \citet{sompolinsky1999svm} calculated learning curves for support vector machines, but not kernel regression, in the limit of number of training samples going to infinity for dot product kernels with binary inputs using a replica method. Our theory applies to general kernels and finite size datasets. In the infinite training set limit, they observed several learning stages where each spectral mode is learned with a different rate. We observe similar phenomena in kernel regression. In a similar spirit, \cite{cohen2019learning} calculates learning curves for infinite-width neural networks using a path integral formulation and a replica analysis but does not discuss the spectral dependence of the generalization error.

In the infinite width limit, neural networks have many more parameters than training samples yet they do not overfit \cite{ZHANGDeep}. Some authors suggested that this is a consequence of the training procedure since stochastic gradient descent is implicitly biased towards choosing the simplest functions that interpolate the training data \cite{belkin2018reconciling,belkin2018understand, xu2019frequency, jacot2018neural}. Other studies have shown that neural networks fit the low frequency components of the target before the high frequency components during training with gradient descent \cite{xu2018training,rahaman2018spectral,zhang2019explicitizing,luo2019theory}. 
In addition to training dynamics, recent works such as \cite{yang2019finegrained, bietti2019inductive, cao2019understanding} have discussed how the spectrum of kernels impacts its smoothness and approximation properties. Here we explore similar ideas by explicitly calculating average case learning curves for kernel regression and studying its dependence on the kernel's eigenspectrum.




\section{Kernel Regression Learning Curves}

We start with a general theory of kernel regression. Implications of our theory for dot product kernels including NTK and trained neural networks are described in Section \ref{SecDot}.

\subsection{Notation and Problem Setup}

We start by defining our notation and setting up our problem. Our initial goal is to derive a mathematical expression for generalization error in kernel regression, which we will analyze in the subsequent sections using techniques from the Gaussian process literature \cite{sollich1999learning,sollich2001mismatch,sollich2002approx} and statistical physics \cite{sherrington_spinglass,mezard1987spin}. 

The goal of kernel regression is to learn a function $f: \mathcal{X} \to \mathbb{R}^C$ from a finite number of observations \cite{Wahba90a,scholkopf_smola}. In developing our theory, we will first focus on the case where $C=1$, and later extend our results to $C>1$ as we discuss in Section \ref{Multiple_Classes}.   
Let $\{\mathbf{x}_i,y_i\} \in \mathcal{X}\times \mathbb{R}$, where $ \mathcal{X}\subseteq \mathbb{R}^{d}$, be one of the $p$ training examples and let $\mathcal{H}$ be a Reproducing Kernel Hilbert space (RKHS) with inner product $\braket{\cdot, \cdot}_{\mathcal{H}}$. To avoid confusion with our notation for averaging, we will always decorate angular brackets for Hilbert inner product with $\mathcal{H}$ and a comma. Kernel ridge regression is defined as: 
\begin{equation}\label{Kreg}
    \min_{f\in\mathcal{H}} \sum_{i=1}^p ( f(\mathbf{x}_i) - y_i)^2 + \lambda ||f||_{\mathcal{H}}^2.
\end{equation}
The $\lambda \to 0$ limit is referred to as interpolating kernel regression, and, as we will discuss later, relevant to training wide neural networks. The unique minimum of the convex optimization problem is given by
\begin{equation}\label{minfun}
    f(\mathbf{x}) = \mathbf{y}^\top (\mathbf{K}+\lambda \mathbf{I})^{-1} \mathbf{k}(\mathbf{x}),
\end{equation}
where $K(\cdot, \cdot)$ is the reproducing kernel for $\mathcal{H}$, $\mathbf{K}$ is the $p \times p$ kernel gram matrix $K_{ij} = K(\mathbf{x}_i,\mathbf{x}_j)$, and $k(\mathbf{x})_i = K(\mathbf{x}, \mathbf{x}_i)$. Lastly, $\mathbf{y} \in \mathbb{R}^p$ is the vector of target values ${y}_i = f^*(\mathbf{x}_i)$. For interpolating kernel regression, when the kernel is invertible, the solution is the same except that $\lambda = 0$, meaning that training data is fit perfectly. The proof of this optimal solution is provided in the Supplementary Information (SI) Section \ref{SIKernelMachines}. 

Let $p(\mathbf{x})$ be the probability density function from which the input data are sampled. The generalization error is defined as the expected risk with expectation taken over new test points sampled from the same density $p(\mathbf{x})$. For a given dataset $\{\mathbf{x}_i\}$ and target function $f^*(\mathbf{x})$, let $f_K(\mathbf{x}; \{\mathbf{x}_i\}, f^*)$ represent the function learned with kernel regression. The generalization error for this dataset and target function is
\begin{equation}
    E_g(\{\mathbf{x}_i\}, f^*) = \int d\mathbf{x}\,p(\mathbf{x}) \left(f_K(\mathbf{x}; \{\mathbf{x}_i\}, f^*) - f^*(\mathbf{x})  \right)^2. 
\end{equation}
To calculate the \textit{average case} performance of kernel regression, we average this generalization error over the possible datasets $\{\mathbf{x}_i \}$ and target functions $f^*$
\begin{align}\label{eq:Eg}
   E_g &= \left< E_g(\{\mathbf{x}_i\}, f^*) \right>_{\{\mathbf{x}_i\}, f^*}.
\end{align}
Our aim is to calculate $E_g$ for a general kernel and a general distribution over teacher functions. 

For our theory, we will find it convenient to work with the feature map defined by the Mercer decomposition. By Mercer's theorem \cite{mercer1909xvi,GPMLRasmussen} 
, the kernel admits a representation in terms of its $M$ kernel eigenfunctions $\{ \phi_\rho(\mathbf{x}) \}$, 
\begin{equation}
    K(\mathbf{x}, \mathbf{x}') = \sum_{\rho = 1}^M \lambda_\rho \phi_{\rho}(\mathbf{x}) \phi_\rho(\mathbf{x}') = \sum_{\rho=1}^M \psi_\rho(\mathbf{x}) \psi_\rho(\mathbf{x}_i),
\end{equation}
where $\psi_\rho(\mathbf{x}) = \sqrt{\lambda_\rho} \phi(\mathbf{x})$ is the feature map we will work with. In our analysis, $M$ will be taken to be infinite, but for the derivation of the learning curves, we will first consider $M$ as a finite integer. The eigenfunctions and eigenvalues are defined with respect to the probability measure that generates the data $d\mu(\mathbf{x}) = p(\mathbf{x}) d\mathbf{x}$
\begin{equation}\label{eq:eigenvalue}
    \int  d \mathbf{x}'\,p(\mathbf{x}') K(\mathbf{x}, \mathbf{x}') \phi_\rho(\mathbf{x}')  = \lambda_\rho \phi_\rho(\mathbf{x}).
\end{equation}

We will also find it convenient to work with a vector representation of the RKHS functions in the feature space. Kernel eigenfunctions form a complete orthonormal basis, allowing the expansion of the target function $f^*$ and learned function $f$ in terms of features $\{ \psi_\rho(\mathbf{x}) \}$  
\begin{align}
     f^*(\mathbf{x}) = \sum_\rho \overline{w}_\rho \psi_\rho(\mathbf{x}), \quad
    f(\mathbf{x}) = \sum_\rho w_\rho \psi_\rho(\mathbf{x}).
\end{align}
Hence, $M$-dimensional vectors $\mathbf{w}$ and $\mathbf{\overline{w}}$ constitute a representation of $f$ and $f^*$ respectively in the feature space. 

We can also obtain a feature space expression for the optimal kernel regression function \eqref{minfun}. Let $\mathbf{\Psi} \in \mathbb{R}^{M \times p}$ be feature matrix for the sample so that $\mathbf{\Psi}_{\rho, i} = \psi_\rho(\mathbf{x}_i)$. With this representation, kernel ridge regression \eqref{Kreg} can be recast as the optimization problem $\min_{\mathbf{w} \in \mathbb{R}^M, \,\Vert \mathbf{w}\Vert_2 < \infty} \Vert \mathbf{\Psi}^\top \mathbf{w} - \mathbf{y}\Vert^2 + \lambda \Vert \mathbf{w} \Vert^2$,
whose solution is
\begin{align}\label{wmin}
    \mathbf{w} = (\mathbf{\Psi} \mathbf{\Psi}^\top + \lambda \mathbf{I})^{-1} \mathbf{\Psi} \mathbf{y}.
\end{align}

Another novelty of our theory is the decomposition of the generalization error into its contributions from different eigenmodes. The feature space expression of the generalization error after averaging over the data distribution can be written as:
\begin{align}
E_g = \sum_{\rho}E_{\rho}, \quad E_{\rho} \equiv \lambda_\rho \left< (w_\rho - \overline{w}_\rho)^2 \right>_{\{\mathbf{x}_i\}, \mathbf{\overline w}},
\end{align}
where we identify $E_\rho$ as the generalization error in mode $\rho$.
 \begin{proof}
 \begin{align}
     \nonumber
     E_g &= \left<(f(x) - f^*(x))^2\right>_{\mathbf{x}, \{ \mathbf{x}_i\}, f*}
     \\
     \nonumber
     &= \sum_{\rho,\gamma} \left< (w_\rho - \overline{w}_\rho)(w_\gamma -\overline{w}_\gamma) \right>_{\{ \mathbf{x}_i\}, f* } \left< \psi_\rho(\mathbf{x}) \psi_\gamma(\mathbf{x}) \right>_{\mathbf{x}}
     \\
     &= \sum_\rho \lambda_\rho  \left< (w_\rho - \overline{w}_\rho)^2 \right>_{\{\mathbf{x}_i\}, \mathbf{\overline w}} = \sum_\rho E_\rho.
 \end{align}
 \end{proof}
%

 We introduce a matrix notation for RKHS eigenvalues $\mathbf{\Lambda}_{\rho,\gamma} \equiv \delta_{\rho,\gamma} \lambda_\rho$ for convenience. Finally, with our notation set up, we can present our first result about generalization error.

\begin{proposition}\label{prop1}
For the $\mathbf{w}$  that minimizes the training error (eq. \eqref{wmin}), 
the generalization error (eq. \eqref{eq:Eg}) is given by
\begin{align}\label{mainEg}
    E_g = {\rm Tr} \left(\mathbf{D} \left< \mathbf{G}^2 \right>_{\{\mathbf{x}_i \}} \right),
\end{align}
which can be decomposed into modal generalization errors
\begin{align}\label{mainErho}
    E_\rho = \sum_{\gamma} \mathbf{D}_{\rho,\gamma} \left< \mathbf{G}^2_{\gamma,\rho} \right>_{\{\mathbf{x}_i\}},
\end{align}
where 
\begin{align}
    \mathbf{G} = \left (\frac{1}{\lambda} \mathbf{\Phi} \mathbf{\Phi}^\top + \mathbf{\Lambda}^{-1} \right)^{-1},\quad  \mathbf{\Phi} = \mathbf{\Lambda}^{-1/2} \mathbf{\Psi}.
\end{align}
 and 
 \begin{equation}
    \mathbf{D} = \mathbf{\Lambda}^{-1/2} \left< \mathbf{\overline{w}}\mathbf{\overline{w}}^\top \right>_{\mathbf{\overline{w}}} \mathbf{\Lambda}^{-1/2}.
\end{equation}
\end{proposition}

We leave the proof to SI Section \ref{SIDerivGenErr} but provide a few cursory observations of this result. First, note that all of the dependence on the teacher function comes in the matrix $\mathbf{D}$ whereas all of the dependence on the empirical samples is in $\mathbf{G}$. In the rest of the paper, we will develop multiple theoretical methods to calculate the generalization error given by expression \eqref{mainEg}. 

Averaging over the target weights in the expression for $\mathbf{D}$ is easily done for generic weight distributions. The case of a fixed target is included by choosing a delta-function distribution over $\mathbf{\overline{w}}$. 

We present two methods for computing the nontrivial average of the matrix $\mathbf{G}^2$ over the training samples $\{\mathbf{x}_i\}$. First, we consider the effect of adding a single new sample to $\mathbf{G}$ to derive a recurrence relation for $\mathbf{G}$ at different number of data points. This method generates a partial differential equation that must be solved to compute the generalization error.  Second, we use a replica method and a saddle point approximation to calculate the matrix elements of $\mathbf{G}$. These approaches give identical predictions for the learning curves of kernel machines.

For notational simplicity, in the rest of the paper, we will use $\braket{\ldots}$ to mean $\braket{\ldots}_{{\{\mathbf{x}_i\}, \mathbf{\overline w}}}$ unless stated otherwise. In all cases, the quantity inside the brackets will depend either on the data distribution or the distribution of target weights, but not both. 

\subsection{Continuous Approximation to Learning Curves}

First, we adopt a method following  \citet{sollich1999learning,sollich2001mismatch} and \citet{sollich2002approx}  to calculate the generalization error. We generalize the definition of $\mathbf{G}$ by introducing an auxiliary parameter $v$, and make explicit its dataset size, $p$, dependence: 
\begin{equation}
    \mathbf{\tilde G}(p,v) = \left(\frac{1}{\lambda} \mathbf{\Phi} \mathbf{\Phi}^\top + \mathbf{\Lambda}^{-1} + v \mathbf{I} \right)^{-1}.
\end{equation}
Note that the quantity we want to calculate is given by
\begin{align}\label{mainRel}
\left< \mathbf{G}^2(p) \right> = \left.- \frac{\partial}{\partial v} \left< \mathbf{\tilde G}(p,v) \right>\right|_{v=0}.
\end{align} 

By considering the effect of adding a single randomly sampled input $\mathbf{x'}$, and treating $p$ as a continuous parameter, we can derive an approximate quasi-linear partial differential equation (PDE) for the average elements of $\mathbf{G}$ as a function of the number of data points $p$ (see below for a derivation):
\begin{equation}\label{pde}
    \frac{\partial \left<\mathbf{\tilde{G}}(p,v)\right>}{\partial p}  =  \frac{1}{\lambda + \text{Tr} \left< \mathbf{\tilde{G}}(p,v) \right>} \frac{\partial}{\partial v} \left< \mathbf{\tilde{G}}(p,v) \right>,
\end{equation}
with the initial condition $\mathbf{\tilde{G}}(0,v) = (\mathbf{\Lambda}^{-1} + v \mathbf{I})^{-1}$, which follows from  $\mathbf{\Phi} \mathbf{\Phi}^\top = \bm{0}$ when there is no data.  
Since $\mathbf{\tilde{G}}$ is initialized as a diagonal matrix, the off-diagonal elements will not vary under the dynamics and $\left< \mathbf{\tilde{G}}(p,v) \right>$ will remain diagonal for all $(p,v)$. We will use the solutions to this PDE and relation \eqref{mainRel} to arrive at an approximate expression for the generalization error $E_g$ and the mode errors $E_\rho$.

\begin{proof}[Derivation of the PDE approximation \eqref{pde}] 
Let $\phi \in \mathbb{R}^M$ represent the new feature to be added to $\mathbf{G}^{-1}$ so that $\phi_\rho = \phi_\rho(\mathbf{x}')$ where $\mathbf{x}' \sim p(\mathbf{x}')$ is a random sample from the data distribution. Let $\left< \mathbf{\tilde{G}}(p,v)\right>_{\mathbf{\Phi}}$ denote the matrix $\mathbf{\tilde{G}}$ averaged over it's $p$-sample design matrix $\mathbf{\Phi}$. By the Woodbury matrix inversion formula 
\begin{align}
   & \left< \mathbf{\tilde{G}}(p+1,v) \right>_{\mathbf{\Phi}, \phi} = \left< \left( \mathbf{\tilde{G}}(p,v)^{-1} + \frac{1}{\lambda} \phi \phi^\top \right)^{-1} \right>_{\mathbf{\Phi}, \phi}
    \nonumber \\
    &\quad= \left< \mathbf{\tilde{G}}(p,v) \right>_{\mathbf{\Phi}} - \left< \frac{\mathbf{\tilde{G}}(p,v) \phi \phi^\top \mathbf{\tilde{G}}(p,v)}{\lambda + \phi^\top \mathbf{\tilde{G}}(p,v) \phi} \right>_{\mathbf{\Phi}, \phi}.
\end{align}
Performing the average of the last term on the right hand side is  difficult so we resort to an approximation, where the numerator and denominator are averaged separately. 
\begin{align}
    \left< \mathbf{\tilde{G}}(p+1,v) \right>_{\mathbf{\Phi}, \phi} &\approx 
   \left< \mathbf{\tilde{G}}(p,v) \right>_{\mathbf{\Phi}} - \frac{\left< \mathbf{\tilde{G}}(p,v)^2 \right>_{\mathbf{\Phi}}}{\lambda + \text{Tr} \left< \mathbf{\tilde{G}}(p,v)\right>_{\mathbf{\Phi}}},
\end{align}
where we used the fact that $\left< \phi_\rho(\mathbf{x'}) \phi_\gamma(\mathbf{x}')\right>_{\mathbf{x}' \sim p(\mathbf{x}')} = \delta_{\rho,\gamma}$.

Treating $p$ as a continuous variable and taking a continuum limit of the finite differences given above, we arrive at \eqref{pde}.
\end{proof}

Next, we present the solution to the PDE \eqref{pde} and the resulting generalization error.

\begin{proposition}\label{prop2}
Let $g_\rho(p,v)  = \left< \mathbf{\tilde{G}}(p,v)_{\rho \rho} \right>$ represent the diagonal elements of the average matrix $\left< \mathbf{\tilde{G}}(p,v) \right>$. These matrix elements satisfy the implicit relationship
\begin{equation}
    g_\rho(p,v)= \left(\frac{1}{\lambda_\rho} + v + \frac{p}{\lambda + \sum_{\gamma=1}^M g_\gamma(p,v)}\right)^{-1}.
\end{equation}
\end{proposition}

This implicit solution is obtained from the method of characteristics which we provide in Section \ref{SISolutionPDE} of the SI. 

\begin{proposition}\label{prop3}
Under the PDE approximation \eqref{pde}, the average error $E_{\rho}$ associated with mode $\rho$ is 
\begin{equation}\label{modeError}
    E_\rho(p) = \frac{\braket{\overline{w}_\rho^2}}{\lambda_\rho}\left(\frac{1}{\lambda_\rho} + \frac{p}{\lambda+t(p)}\right)^{-2} \left(1 - \frac{p \gamma(p)}{(\lambda + t(p))^2}\right)^{-1},
\end{equation}
where $t(p)\equiv \sum_\rho g_\rho(p,0)$ is the solution to the implicit equation 
\begin{equation}\label{t_func}
    t(p) = \sum_\rho \left(\frac{1}{\lambda_\rho} + \frac{p}{\lambda+t(p)}\right)^{-1},
\end{equation}
and $\gamma(p)$ is defined as 
\begin{equation}
    \gamma(p) = \sum_\rho \left(\frac{1}{\lambda_\rho} + \frac{p}{\lambda+t(p)}\right)^{-2}.
\end{equation}
%
\end{proposition}

The full proof of this proposition is provided in Section \ref{SISolutionPDE} of the SI. We show the steps required to compute theoretical learning curves numerically in Algorithm \ref{alg:learning_curve}.

\begin{algorithm}[h]
   \caption{Computing Theoretical Learning Curves}
   \label{alg:learning_curve}
\begin{algorithmic}
   \STATE {\bfseries Input:} RKHS spectrum $\{\lambda_\rho\}$, target function weights $\{\overline{w}_\rho\}$, regularizer $\lambda$, sample sizes $\{p_i\}$, $i=1,...,m$;
   \FOR{$i=1$ {\bfseries to} $m$}

   \STATE Solve numerically $t_i = \sum_\rho \left(\frac{1}{\lambda_\rho} + \frac{p_i}{\lambda + t_i}\right)^{-1}$
   \STATE Compute $\gamma_i = \sum_\rho \left(\frac{1}{\lambda_\rho} + \frac{p_i}{\lambda + t_i}\right)^{-2}$
   \STATE $E_{\rho,i} = \frac{\braket{\overline{w}_\rho^2}}{\lambda_\rho}  \left(\frac{1}{\lambda_\rho} + \frac{p_i}{\lambda + t_i}\right)^{-2} \left(  1 - \frac{p_i \gamma_i}{(\lambda+t_i)^2}  \right)^{-1}$
   \ENDFOR
\end{algorithmic}
\end{algorithm}

In eq. \eqref{modeError}, the target function sets the overall scale of $E_{\rho}$. That $E_{\rho}$ depends only on $\bar w_{\rho}$, but not other target modes, is an artifact of our approximation scheme, and in a full treatment may not necessarily hold. The spectrum of the kernel affects all modes in a nontrivial way. When we apply this theory to neural networks in Section \ref{SecDot}, the information about the architecture of the network will be in the spectrum $\{\lambda_{\rho}\}$. The dependence on number of samples $p$ is also nontrivial, but we will consider various informative limits below. 

We note that though the mode errors fall asymptotically like $p^{-2}$ (SI Section \ref{SIPowerLaw}), the total generalization error $E_g$ can scale with $p$ in a nontrivial manner. For instance, if $\overline{w}^2_{\rho} \lambda_{\rho} \sim {\rho}^{-a}$ and $\lambda_{\rho} \sim {\rho}^{-b}$ then a simple computation (SI Section \ref{SIPowerLaw}) shows that $E_g \sim p^{- \min\{a-1,2b\}}$ as $p\to\infty$ for ridgeless regression and $E_g \sim p^{- \min\{ a-1,2b\}/b}$ for explicitly regularized regression. This is consistent with recent observations that total generalization error for neural networks and kernel regression falls in a power law $E_g \sim p^{-\beta}$ with $\beta$ dependent on kernel and target function \cite{hestness2017deep,spigler2019asymptotic}.




\subsection{Computing Learning Curves with Replica Method}\label{sec:main_replica}
The result of the continuous approximation can be obtained using another approximation method, which we outline here and detail in SI Section \ref{SIReplicaCalc}. We perform the average of matrix $\mathbf{G}(p,v)$ over the training data, using the replica method \cite{sherrington_spinglass,mezard1987spin} from statistical physics and a finite size saddle-point approximation, and obtain identical learning curves to Proposition \ref{prop3}. Our starting point is a Gaussian integral representation of the matrix inverse
\begin{align}
    \left< \mathbf{G}(p,v)_{\rho,\gamma} \right> &= \frac{\partial^2}{\partial h_\rho \partial h_\gamma} R(p,v,\mathbf{h}) |_{\mathbf{h}=0}, \nonumber
    \\
    R(p,v,\mathbf{h}) &\equiv \left< \frac{1}{Z} \int d\mathbf{u} \ e^{-\frac{1}{2} \mathbf{u}^\top ( \frac{1}{\lambda} \mathbf{\Phi} \mathbf{\Phi}^\top + \mathbf{\Lambda}^{-1} + v \mathbf{I} ) \mathbf{u} + \mathbf{h} \cdot \mathbf{u}}  \right>,
\end{align}
where  $Z = \int d\mathbf{u} \ e^{-\frac{1}{2} \mathbf{u}^\top ( \frac{1}{\lambda} \mathbf{\Phi} \mathbf{\Phi}^\top + \mathbf{\Lambda}^{-1} + v \mathbf{I} ) \mathbf{u}}$. Since $Z$ also depends on the dataset (quenched disorder) $\mathbf{\Phi}$,  to make the average over $\mathbf{\Phi}$ tractable, we use the following limiting procedure: $Z^{-1} = \lim_{n \to 0} Z^{n-1}$. As is common in the physics of disordered systems \cite{mezard1987spin}, we compute $R(p,v,\mathbf{h})$ for integer $n$ and analytically continue the expressions in the $n \to 0$ limit under a symmetry ansatz. This procedure produces the same average matrix elements as the continuous approximation discussed in Proposition \ref{prop2}, and therefore the same generalization error given in Proposition \ref{prop3}. Further detail is provided in SI Section \ref{SIReplicaCalc}.

\subsection{Spectral Dependency of Learning Curves}\label{sec:spectral}

We can get insight about the behavior of learning curves by considering ratios between errors in different modes:
\begin{equation}
    \frac{E_\rho}{E_\gamma} = \frac{\braket{\overline{w}_\rho^2}}{\braket{\overline{w}_\gamma^2}}\frac{\lambda_\gamma}{\lambda_\rho} \frac{(\frac{1}{\lambda_\gamma} + \frac{p}{\lambda+t})^2}{(\frac{1}{\lambda_\rho} + \frac{p}{\lambda+t})^2}.
\end{equation}
For small $p$ this ratio approaches $\frac{E_\rho}{E_\gamma} \sim \frac{\lambda_\rho\braket{\overline{w}_\rho^2}}{\lambda_\gamma\braket{\overline{w}_\gamma^2}}$. For large $p$, $\frac{E_\rho}{E_\gamma} \sim \frac{\braket{\overline{w}_\rho^2}/\lambda_\rho}{\braket{\overline{w}_\gamma^2}/\lambda_\gamma}$, indicating that asymptotically ($p \to \infty$), the amount of relative error in mode $\rho$ grows with the ratio $\braket{\overline{w}_\rho^2 }/\lambda_\rho$, showing that the asymptotic mode error is relatively large if the teacher function places large amounts of power in modes that have small RKHS eigenvalues $\lambda_\rho$. 

We can also examine how the RKHS spectrum affects the evolution of the error ratios with $p$. Without loss of generality, we take $\lambda_\gamma > \lambda_\rho$ and 
show in SI Section \ref{SISpectralDependency} that 
\begin{align}\label{dplog} \frac{d}{dp} \log E_\rho > \frac{d}{dp} \log E_\gamma.
\end{align} 
In this sense, the marginal training data point causes a greater percent reduction in generalization error for modes with larger RKHS eigenvalues.

\subsection{Multiple Outputs}\label{Multiple_Classes}
The learning curves we derive for a scalar function can be straightforwardly extended to the case where the function outputs are multivariate: $\mathbf{f}: \mathbb{R}^d \to \mathbb{R}^{C}$. For least squares regression, this case is equivalent to solving $C$ separate learning problems for each component functions $f_c(\mathbf{x})$, $c=1,...,C$. Let 
$\mathbf{y}_c\in\mathbb{R}^{p}$ be the corresponding vectors of target values possibly generated by different target functions, $f^*_c$. The learning problem in this case is
\begin{equation}
    \min_{f\in\mathcal{H}^c} \sum_{c=1}^C \left[ \sum_{i=1}^p (f_c(\mathbf{x}_i) - y_{c,i})^2 + \lambda ||f_c||_{\mathcal{H}}^2 \right].
\end{equation}
%
The solution to the learning problem depends on the same kernel but different targets for each function:
\begin{equation}
    f_c(\mathbf{x}) = \mathbf{y}^\top_c (\mathbf{K} + \lambda \mathbf{I})^{-1} \mathbf{k}(\mathbf{x}), 
    \, c = 1,\ldots,C.
\end{equation}
Our theory can be used to generate predictions for the generalization error of each of the $C$ learned functions, $f_c(\mathbf{x})$, and then summed to obtain the total error. 

\section{Dot Product Kernels on $\mathbb{S}^{d-1}$ and NTK}\label{SecDot}

For the remainder of the paper, we specialize to the case where our inputs are drawn uniformly on $\mathcal{X} = \mathbb{S}^{d-1}$, a $(d-1)$-dimensional unit hyper-sphere. In addition, we will assume that the kernel is a dot product kernel ($K(\mathbf{x},\mathbf{x}') = \kappa(\mathbf{x}^\top \mathbf{x}')$), as is the case for NTK. 
In this setting, the kernel eigenfunctions are spherical harmonics $\{Y_{km}\}$ \cite{bietti2019inductive,costasspherical}, and the Mercer decomposition is given by
\begin{align}\label{dmercer}
    K(\mathbf{x}, \mathbf{x'}) = \sum_{k=0}^\infty \lambda_k \sum_{m=1}^{N(d,k)} Y_{km}(\mathbf{x}) Y_{km}(\mathbf{x'}).
\end{align}
Here, $N(d,k)$ is the dimension of the subspace spanned by $d$-dimensional spherical harmonics of degree $k$. Rotation invariance renders the eigenspectrum degenerate since each of the $N(d,k)$ modes of frequency $k$ share the same eigenvalue $\lambda_k$. A review of these topics is given in SI Sections \ref{SISphericalHarmonics} and \ref{SIDotProductKernels}. 

We briefly comment on another fact that will later be used in our numerical simulations. Dot product kernels admit an expansion in terms of Gegenbauer polynomials $\{Q_k\}$, which form a complete and orthonormal basis for the uniform measure on the sphere \cite{Dai_2013}:  $\kappa(z) = \sum_{k=0}^\infty \lambda_k N(d,k) Q_{k}(z)$. The Gegenbauer polynomials are related to spherical harmonics $\{Y_{km}\}$ through $Q_k(\mathbf{x}^\top \mathbf{x}') = \frac{1}{N(d,k)} \sum_{m=1}^{N(d,k)} Y_{km}(\mathbf{x}) Y_{km}(\mathbf{x}')$ \cite{Dai_2013}  (see SI Sections \ref{SISphericalHarmonics} and \ref{SIDotProductKernels} for a review). 


\subsection{Frequency Dependence of Learning Curves}

In the special case of dot product kernels with monotonically decaying spectra, results given in Section \ref{sec:spectral} indicate that the marginal training data point causes greater reduction in relative error for low frequency modes than for high frequency modes. Monotonic RKHS spectra represent an inductive bias that preferentially favors fitting lower frequencies as more data becomes available. More rapid decay in the spectrum yields a stronger bias to fit low frequencies first.

\begin{figure}[tb]
\centering
\includegraphics[width=0.8
\linewidth]{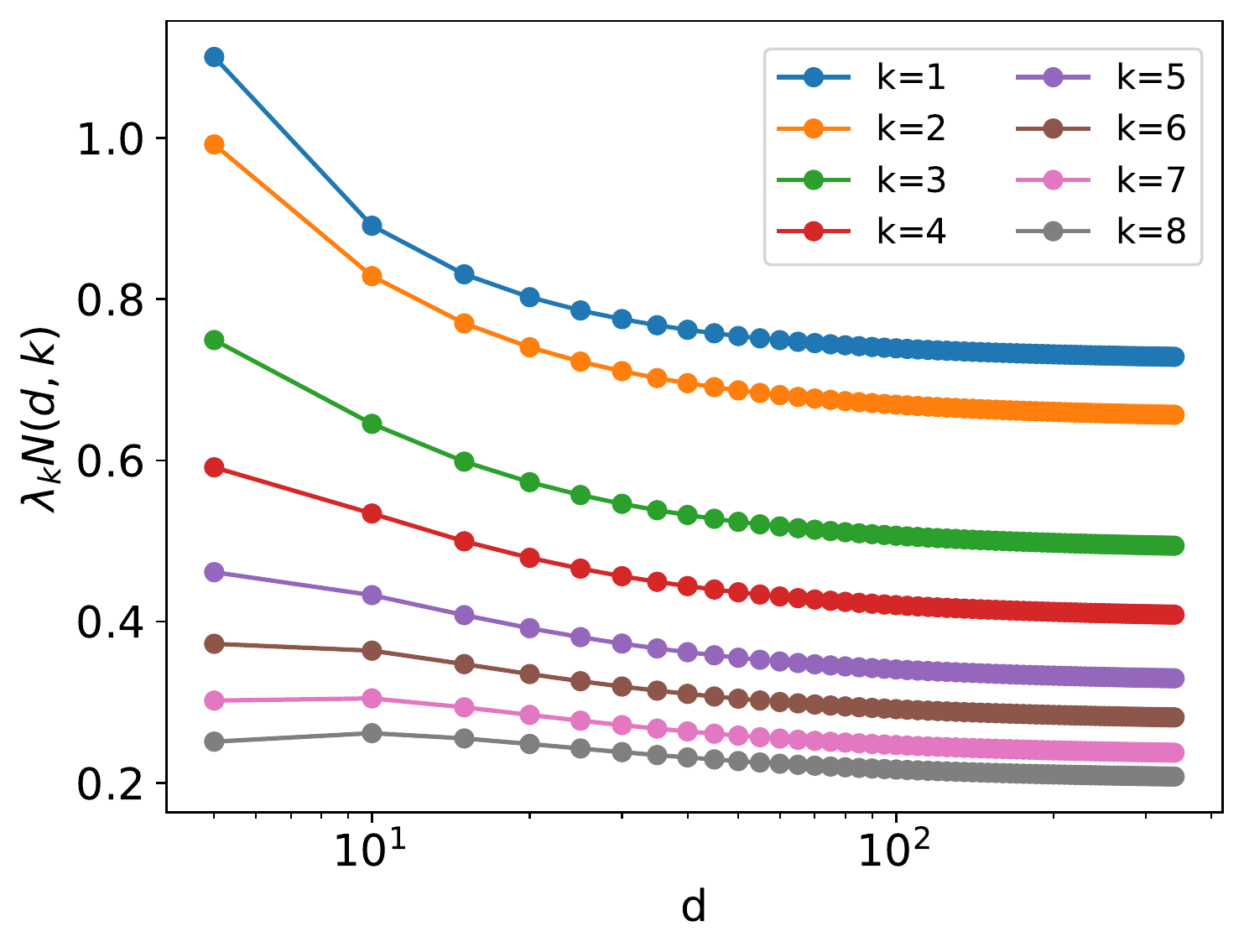}
\caption{\label{fig:NTK_spectrum_dependence} Spectrum of 10-layer NTK multiplied by degeneracy as a function of dimension for various $k$, calculated by numerical integration (SI Section \ref{SIDotProductKernels}). $\lambda_k N(d,k)$ stays constant as input dimension increases, confirming that $\lambda_k N(d,k)^{-1}\sim \mathcal{O}_d(1)$ at large $d$.}
\end{figure}

To make this intuition more precise, we now discuss an informative limit $d\to\infty$ where the degeneracy factor approaches to $N(d,k) \sim {d^k}/{k!}$.  In the following, we replace eigenfunction index $\rho$ with index pair $(k,m)$. Eigenvalues of the kernel scales with $d$ as $\lambda_k \sim N(d,k)^{-1}$ \cite{smola2001dotproduct} in the $d \to \infty$ limit, as we verify numerically in Figure \ref{fig:NTK_spectrum_dependence} for NTK. If 
we take $p=\alpha d^\ell$ for some integer degree $\ell$, then $E_{km}$  exhibits three distinct learning stages. Leaving the details to SI Section \ref{SIFrequencyDependence},
we find that in this limit, for large $\alpha$:
\begin{equation}
\frac{E_{km}(\alpha)}{E_{km}(0)} \approx \left\{\begin{array}{ll}1,& k>\ell \\
 \frac{\text{const.}}{\alpha^2} ,& k=\ell \\
0, & k<\ell\end{array} \right.,
\end{equation}
where the constant is given in SI Section \ref{SIFrequencyDependence}. In other words, $k < l$ modes are perfectly learned, $k = l$ are being learned with an asymptotic $1/\alpha^2$ rate, and  $k > l$ are not learned.    


This simple calculation demonstrates that the lower modes are learned earlier with increasing sample complexity since the higher modes stays stationary until $p$ reaches to the degeneracy of that mode.






\subsection{Neural Tangent Kernel and its Spectrum}

For fully connected architectures, the NTK is a rotation invariant kernel that describes how the predictions of infinitely wide neural networks evolve under gradient flow \cite{jacot2018neural}. Let $\theta_i$ index all of the parameters of the neural network and let $f_\theta(\mathbf{x})$ be the output of the network. Here, we focus on scalar network outputs for simplicity, but generalization to multiple outputs is straightforward, as discussed in Section \ref{Multiple_Classes}. Then the neural tangent kernel is defined as 
\begin{equation}
    K_{\text{NTK}}(\mathbf{x},\mathbf{x}') = \sum_{i} \Big< \frac{\partial f_\theta(\mathbf{x})}{\partial \theta_i} \frac{\partial f_\theta(\mathbf{x}')}{\partial \theta_i} \Big>_{\theta}.
\end{equation}
Let $\mathbf{u}_\theta \in \mathbb{R}^{p}$ be the current predictions of $f_\theta$ on the training data. If the parameters of the model are updated via gradient flow on a quadratic loss, $\frac{d \theta}{dt} = - \nabla_\theta \mathbf{u}_\theta \cdot (\mathbf{u}_\theta - \mathbf{y}) $, then the predictions on the training data evolve with the following dynamics \cite{pehlevan2018flexibility,jacot2018neural,arora2019exact,lee2019wide}
\begin{equation}
\frac{d\mathbf{u}_\theta}{dt}  = -\mathbf{K}_{\text{NTK}} \cdot  \left(\mathbf{u}_\theta - \mathbf{y}\right).
\end{equation}

When the width of the neural network is taken to infinity with proper initialization, where the weights at layer $\ell$ are sampled $W^{(\ell)} \sim \mathcal{N}\left(0,1/ n^{(\ell)}\right)$ where $n^{(\ell)}$ is the number of hidden units in layer $\ell$, the NTK becomes independent of the particular realization of parameters and approaches a deterministic function of the inputs and the nonlinear activation function \cite{jacot2018neural}. Further, the kernel is approximately fixed throughout gradient descent \cite{jacot2018neural,arora2019exact}. If we assume that $\mathbf{u}_\theta = 0 $ at $t=0$, then the final learned function is
\begin{equation}
    f(\mathbf{x}) = \mathbf{y}^\top \mathbf{K}_{\text{NTK}}^{-1} \mathbf{k}(\mathbf{x}).
\end{equation}
Note that this corresponds to ridgeless, interpolating regression where $\lambda = 0$. We will use this correspondence and our kernel regression theory to explain neural network learning curves in the next section. For more information about NTK for fully connected architectures see SI Sections \ref{SINeuralTangent} and \ref{SISpectraNTK}.


To generate theoretical learning curves, we need the eigenspectrum of the kernels involved. For $\mathcal{X}=\mathbb{S}^{d-1}$, it suffices to calculate the projections of the kernel on the Gegenbauer basis $\left < K_{\text{NTK}}(\mathbf{x}),Q_k(\mathbf{x}) \right>_{\mathbf{x}}$, which we evaluate numerically with Gauss-Gegenbauer quadrature (SI Section \ref{SIDotProductKernels}). Further details on NTK spectrum is presented in SI Section \ref{SISpectraNTK}. 






\section{Experiments}

In this section, we test our theoretical results for kernel regression, kernel interpolation and wide networks for various kernels and datasets.

\begin{figure*}
\subfigure[3-layer NTK $d=15$ $\lambda=0$]{\label{fig:kernel_mode_vs_p}\includegraphics[width=0.33\linewidth]{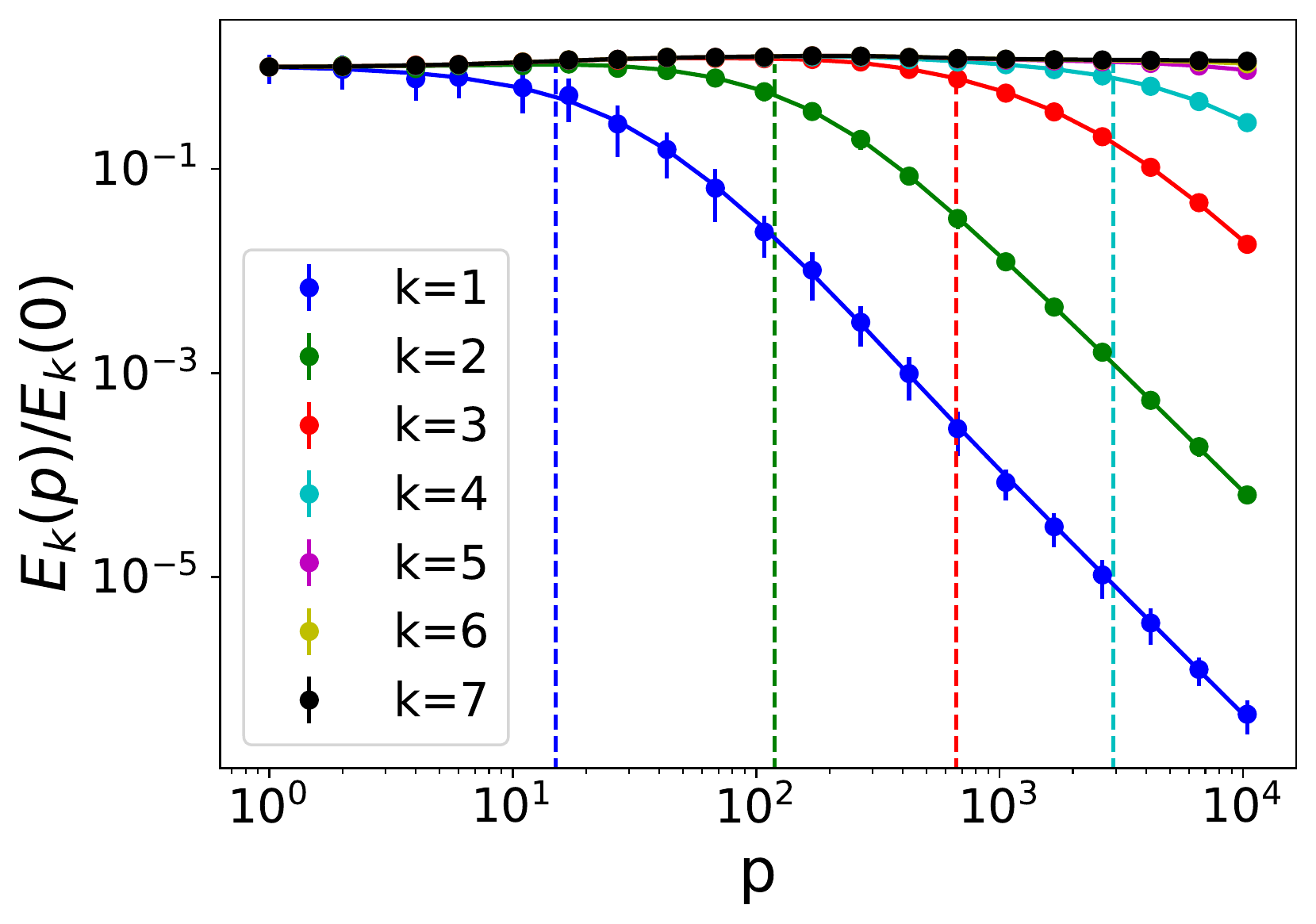}}
\subfigure[3-layer NTK $k=1$ $\lambda=1$]{\label{fig:kernel_mode_vs_dim}\includegraphics[width=0.33\linewidth]{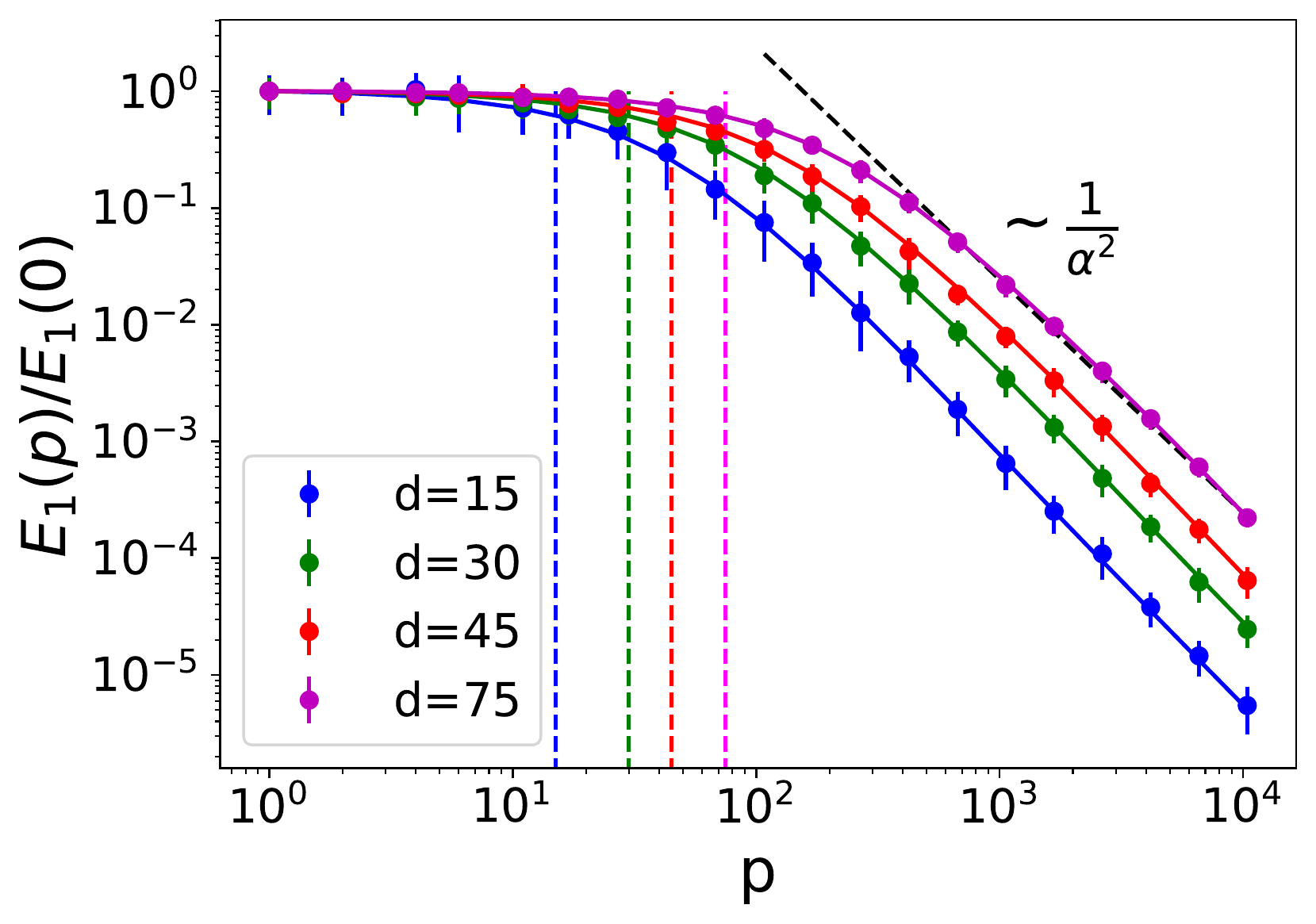}}
\subfigure[10-layer NTK $d=15$ $k=1$]{\label{fig:kernel_mode_vs_lambda}\includegraphics[width=0.33\linewidth]{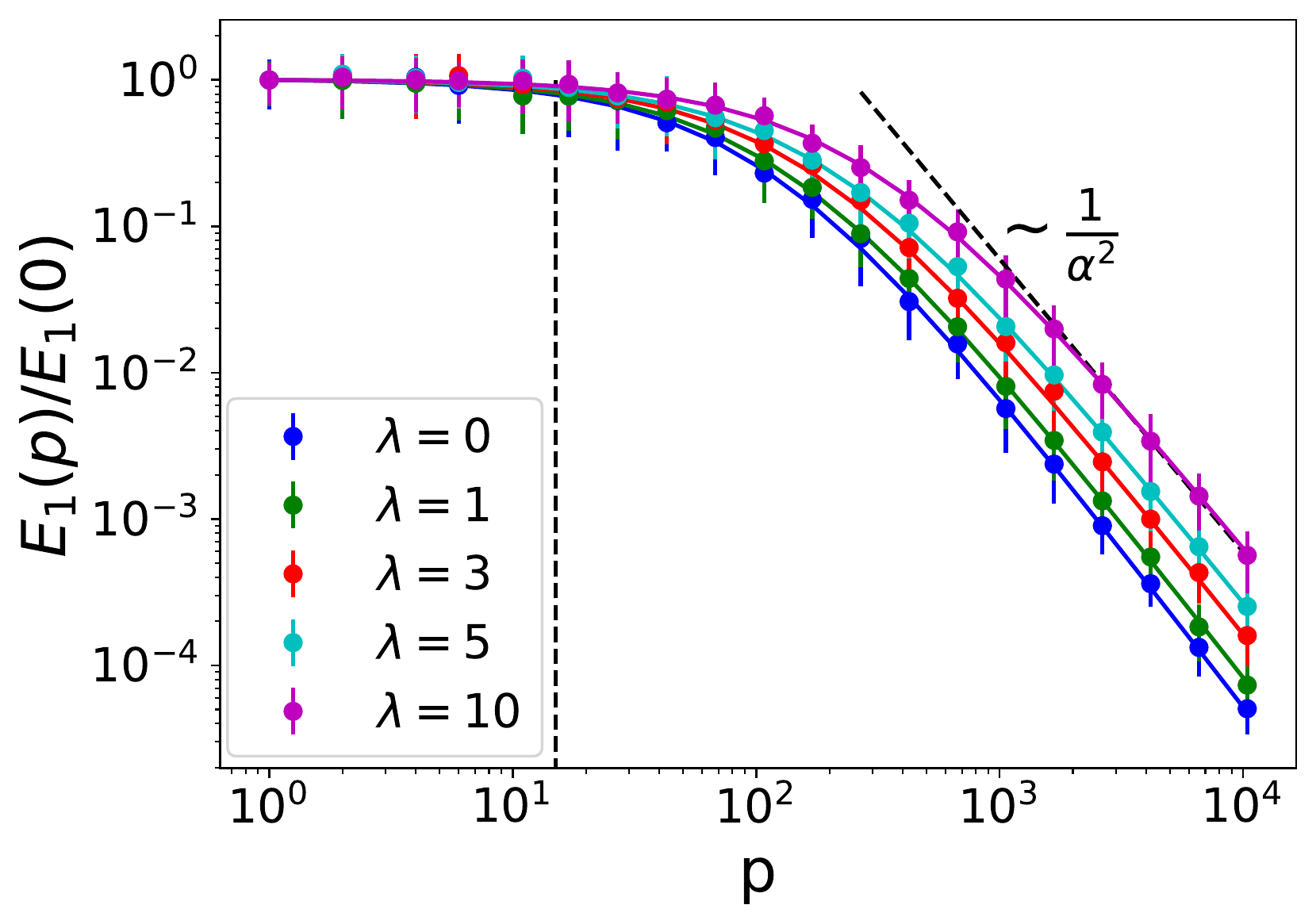}}

\caption{Learning curves for kernel regression with NTK averaged over $50$ trials compared to theory. Error bars are standard deviation. Solid lines are theoretical curves calculated using eq. \eqref{modeError}. Dashed vertical lines indicate the degeneracy $N(d,k)$. (a) Normalized learning curves for different spectral modes. Sequential fitting of mode errors is visible. (b) Normalized learning curves for varying data dimension, $d$. (c) Learning curves for varying regularization parameter, $\lambda$.}
\label{fig:kernel_expts}
\end{figure*}

\begin{figure*}[t]
\subfigure[2-layer NN $N=10000$]{\label{fig:net_a}\includegraphics[width=0.33\linewidth]{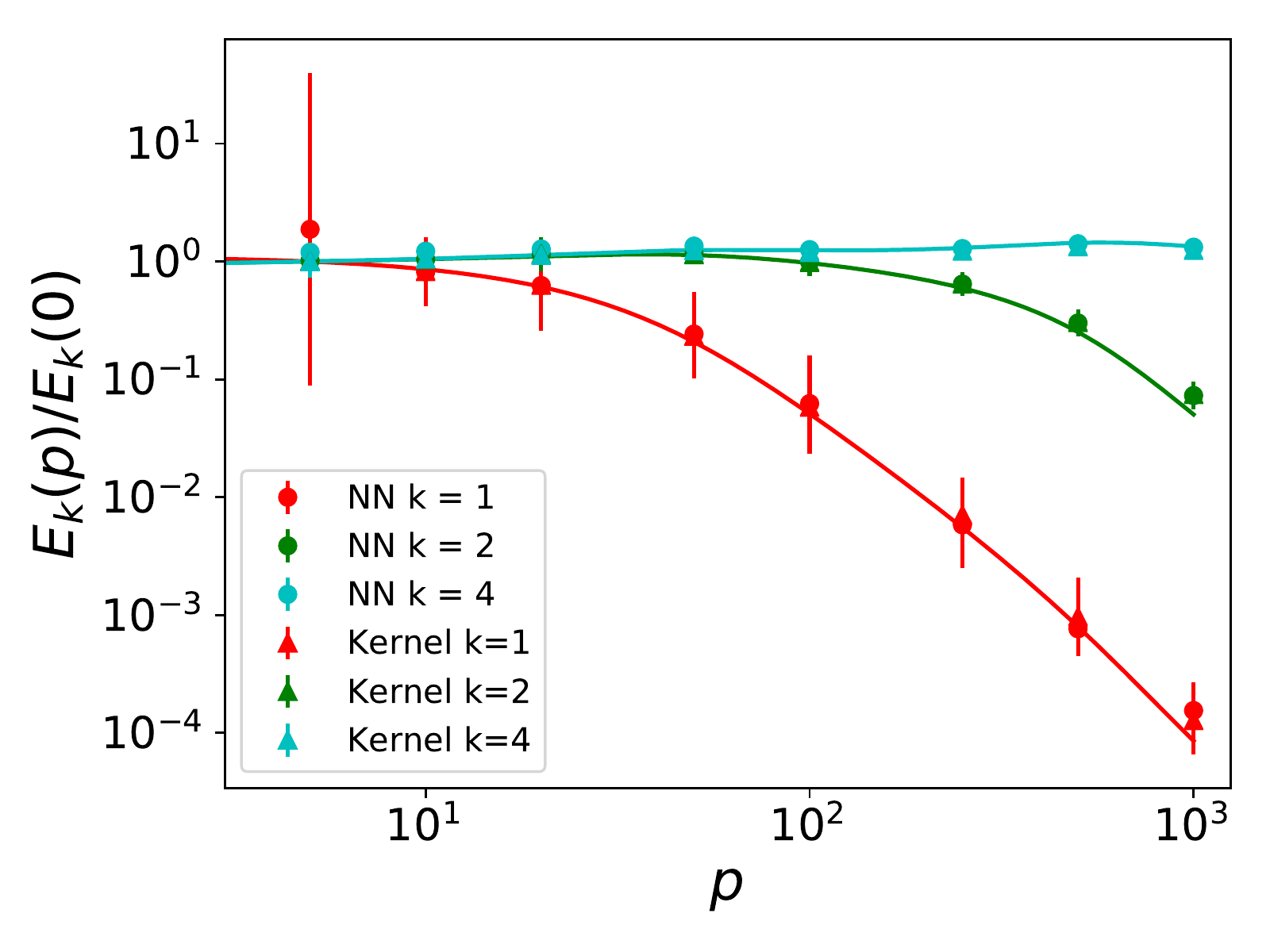}}
\subfigure[4-layer NN $N=500$ ]{\label{fig:net_b}\includegraphics[width=0.33\linewidth]{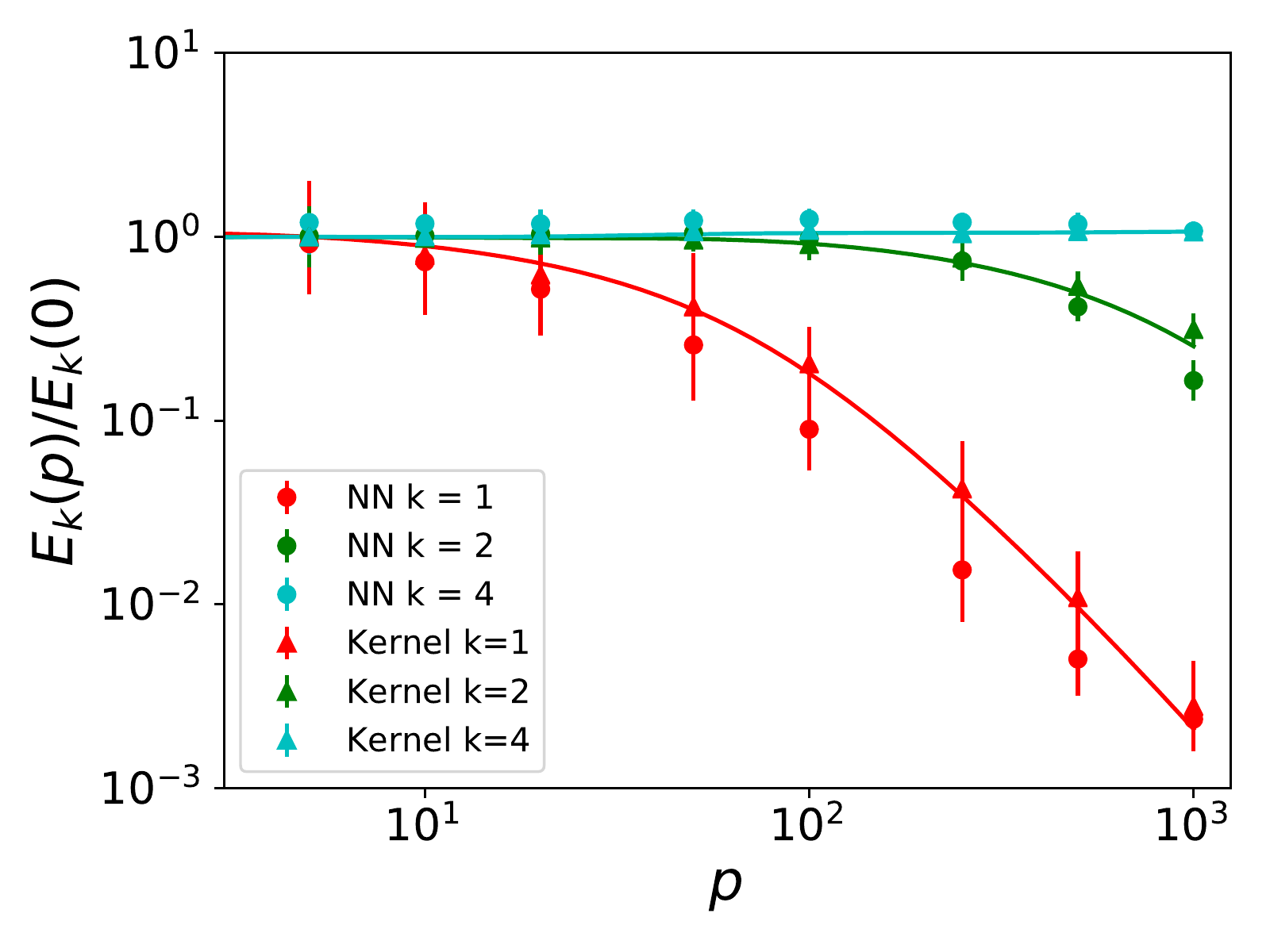}}
\subfigure[2-Layer NN Student-Teacher; $N=8000$ ]{\label{fig:net_c}\includegraphics[width=0.33\linewidth]{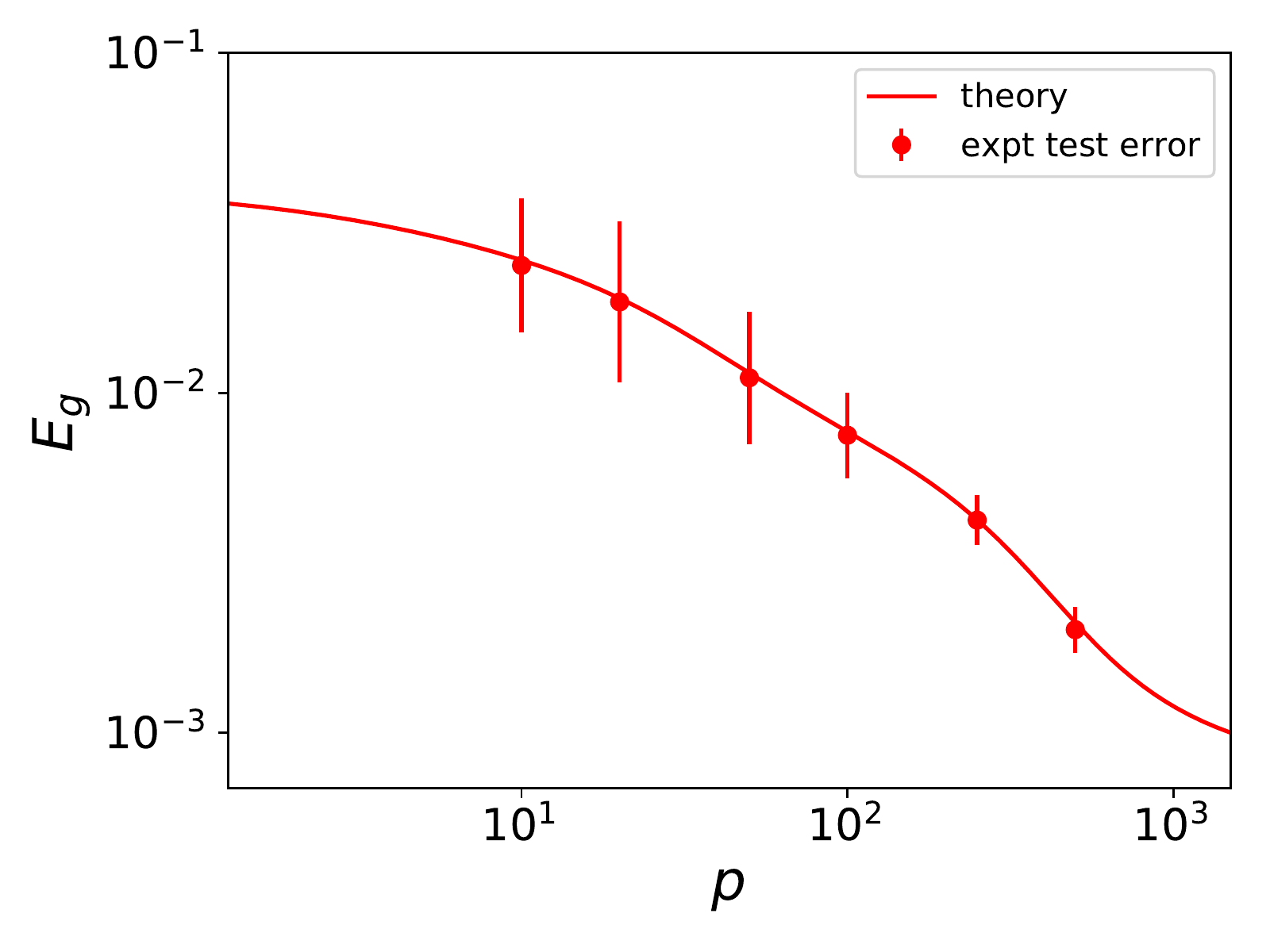}}

\caption{(a) and (b) Learning curves for neural networks (NNs) on ``pure modes" as defined in eq. \eqref{pure_mode}. (c) Learning curve for the student teacher setup defined in \eqref{student_teacher}. The theory curves shown as solid lines are again computed with eq. \eqref{modeError}. The test error for the finite width neural networks and NTK are shown with dots and triangles respectively. The generalization error was estimated by taking a random test sample of $1000$ data points. The average was taken over 25 trials and the standard deviations are shown with errorbars. The networks were initialized with the default Gaussian NTK parameterization \cite{jacot2018neural} and trained with stochastic gradient descent (details in SI Section \ref{SINeuralExperiment}).}
\label{fig:neural_net_expts}
\end{figure*}

\subsection{NTK Regression and Interpolation}\label{sec:reg_expts}

We first test our theory in a kernel regression task with NTK demonstrating the spectral decomposition. In this experiment, the target function is a linear combination of a kernel evaluated at randomly sampled points $\{\mathbf{\overline{x}}_i\}$:
\begin{equation}\label{synthetic_teacher}
    f^*(\mathbf{x}) = \sum_{i=1}^{p'} \overline{\alpha}_i K(\mathbf{x}, \mathbf{\overline{x}}_i),
\end{equation}
where $\overline{\alpha}_i \sim \mathcal{B}(1/2)$ are sampled randomly from a Bernoulli distribution on $\{\pm 1\}$ and $\mathbf{\overline{x}}_i$ are sampled uniformly from $\mathbb{S}^{d-1}$. The points $\mathbf{\overline{x}}_i$ are independent samples from $\mathbb{S}^{d-1}$ and are different than the training set $\{\mathbf{x}_i\}$. The student function is learned with kernel regression by inverting the Gram matrix $\mathbf{K}$ defined on the training samples $\{\mathbf{x}_i\}$ according to eq. \eqref{minfun}. With this choice of target function, exact computation of the mode wise errors $E_k = \sum_{m} E_{km}$ in terms of Gegenbauer polynomials is possible; the formula and its derivation are provided in Section \ref{SIRegressionModeError} of the SI. We compare these experimental mode-errors to those predicted by our theory and find perfect agreement.
For these experiments, both the target and student kernels are taken to be NTK of a 4-layer fully connected ReLU without bias. 

Figure \ref{fig:kernel_expts} shows the errors for each frequency $k$ as a function of sample size $p$. In Figure \ref{fig:kernel_mode_vs_p}, we show that the mode errors sequentially start falling when $p \sim N(d,k)$. Figure \ref{fig:kernel_mode_vs_dim} shows the mode error corresponding to $k=1$ for kernel regression with 3-layer NTK across different dimensions. Higher input dimension causes the frequency modes to be learned at larger $p$. We observe an asymptotic $\sim1/\alpha^2$ decay in modal errors. Finally, we show the effect of regularization on mode errors with a 10-layer NTK in Figure \ref{fig:kernel_mode_vs_lambda}. With increasing $\lambda$, learning begins at larger $p$ values.

\begin{figure*}
    \centering
    \subfigure[Gaussian kernel and measure in $d=20$]{\label{fig:kernel_a}\includegraphics[width=0.33\linewidth]{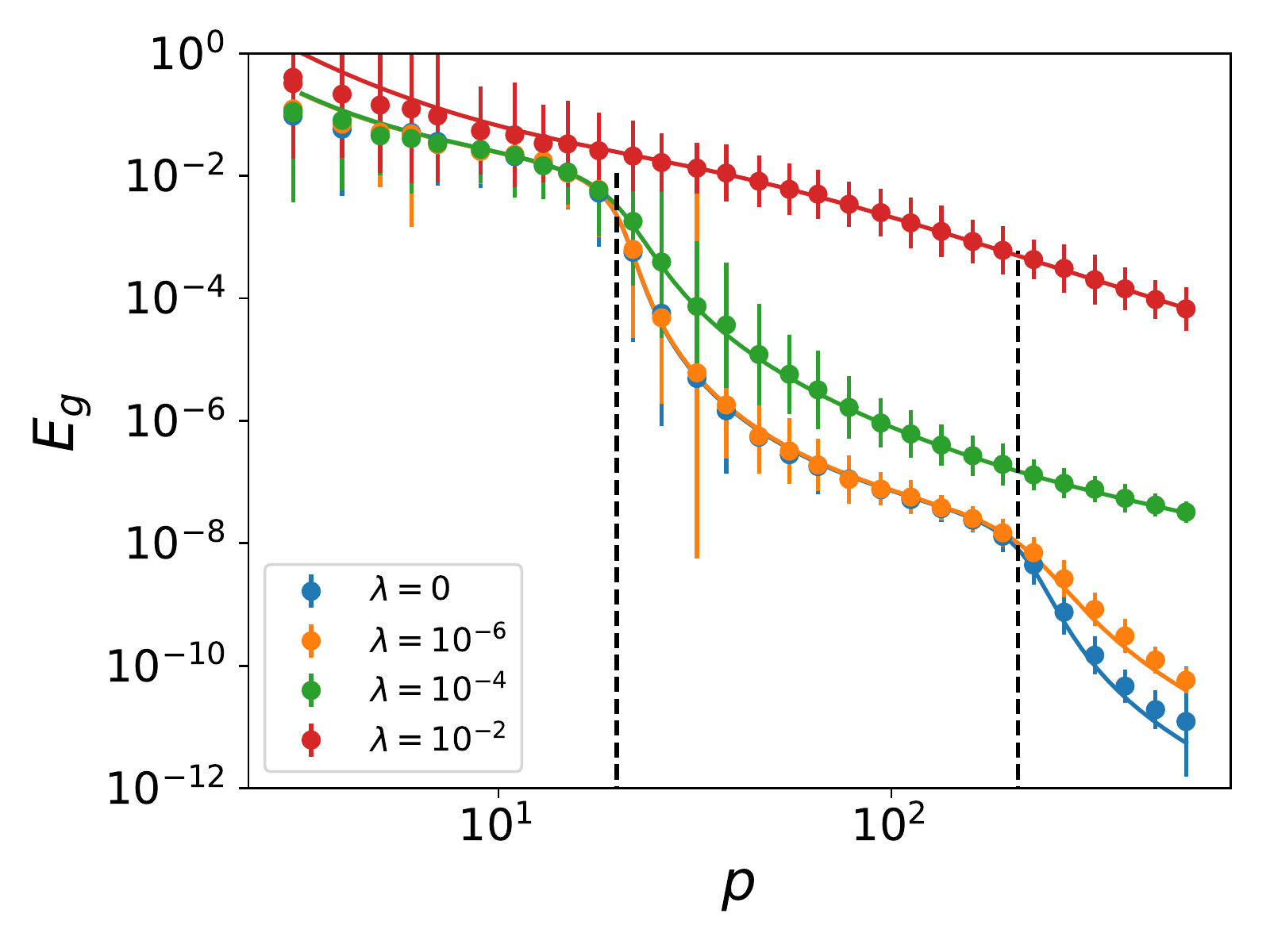}}
    \subfigure[3-Layer NN on MNIST, $N=800$]{\label{fig:kernel_c}\includegraphics[width=0.33\linewidth]{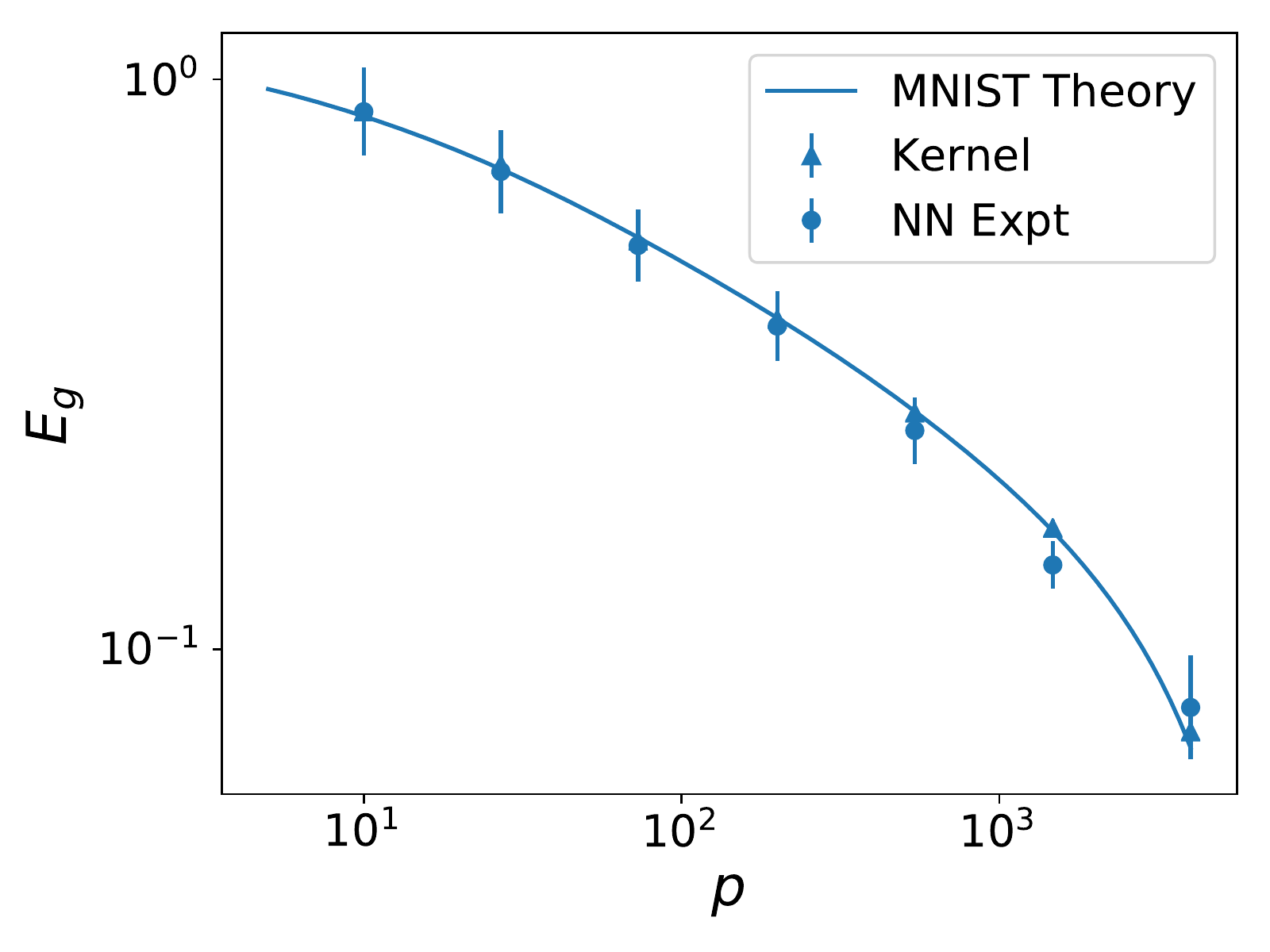}}
    \subfigure[NTK regression on MNIST, $\lambda =0$.]{\label{fig:kernel_b}\includegraphics[width=0.33\linewidth]{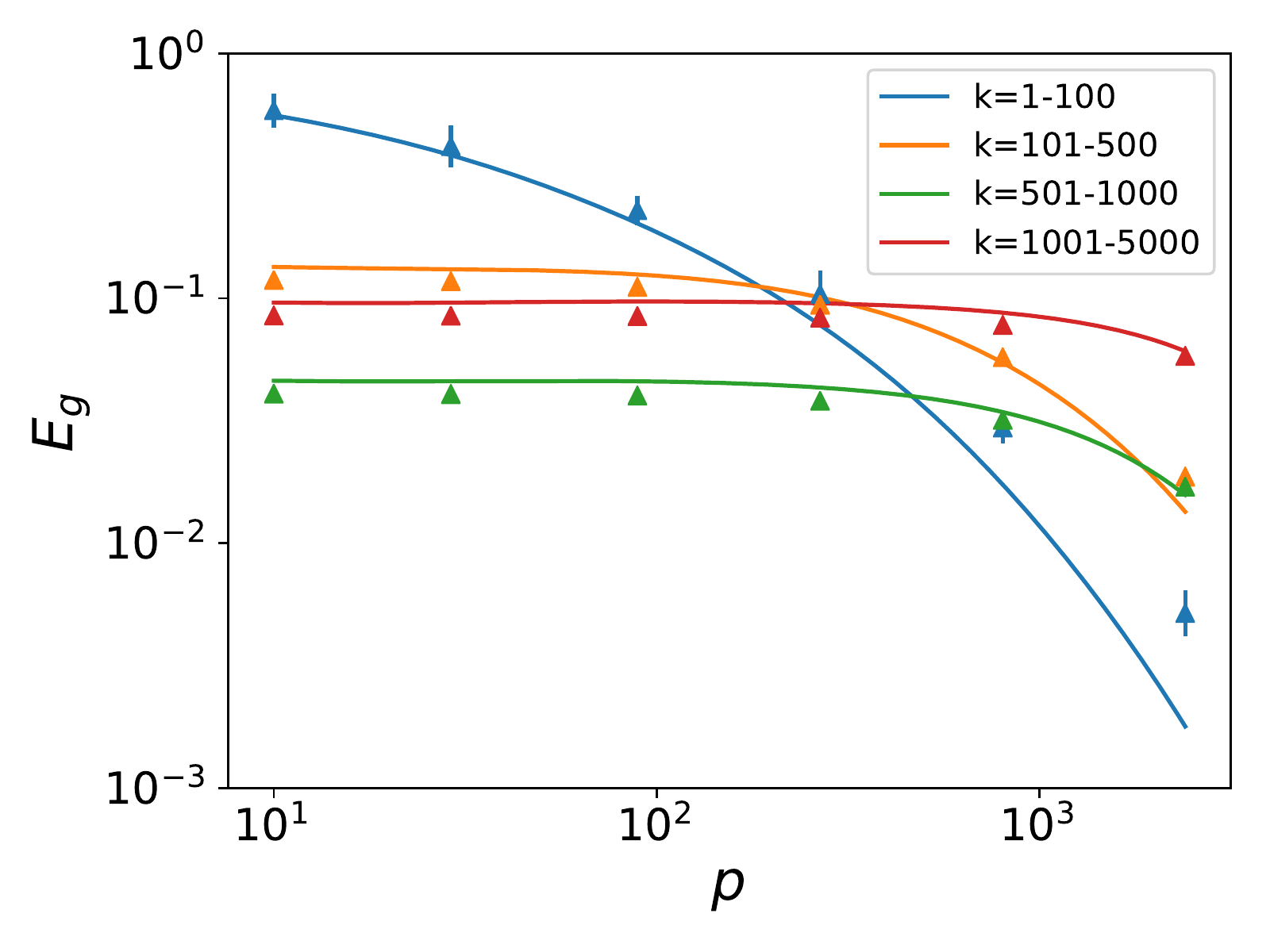}}
    \caption{(a) Learning curves for Gaussian kernel in $d=20$ dimensions with varying $\lambda$. For small $\lambda$, learning stages are visible at $p=N(d,k)$ for $k=1,2$ ($p=20,210$, vertical dashed lines) but the stages are obscured for non-negligible $\lambda$. (b)  Learning curve for 3-layer NTK regression and a neural network (NN) on a subset of 8000 randomly sampled images of handwritten digits from MNIST. (c) Aggregated NTK regression mode errors for the setup in (b).  Eigenmodes of MNIST with larger eigenvalues are learned more rapidly with increasing $p$. 
    \label{fig:RBF}}
\end{figure*}

\subsection{Learning Curves for Finite Width Neural Networks}

Having established that our theory accurately predicts the generalization error of kernel regression with NTK, we now compare the generalization error of finite width neural networks trained on a quadratic loss with the theoretical learning curves for NTK. For these experiments, we use the Neural-Tangents Library \cite{neuraltangents2020} which supports training and inference for both finite and infinite width neural networks.

First, we use ``pure mode" teacher functions, meaning the teacher is composed only of spherical harmonics of the same degree.  For ``pure mode" $k$, the teacher is constructed with the following rule:
\begin{align}\label{pure_mode}
    f^*(\mathbf{x}) = \sum_{i=1}^{p'} \overline{\alpha}_i Q_k(\mathbf{x}^\top \mathbf{\overline{x}}_i),
\end{align}
where again $\overline{\alpha}_i \sim \mathcal{B}(1/2)$ and $\mathbf{\overline{x}}_i \sim p(\mathbf{x})$ are sampled randomly. Figure \ref{fig:net_a} shows the learning curve for a fully connected 2-layer ReLU network with width $N=10000$, input dimension $d=30$ and $p' = 10000$. As before, we see that the lower $k$ pure modes require less data to be fit.  Experimental test errors for kernel regression with NTK on the same synthetic datasets are plotted as triangles. Our theory perfectly fits the experiments.

Results from a 4-layer NN simulation are provided in Figure \ref{fig:net_b}. Each hidden layer had $N=500$ hidden units. We again see that the $k=2$ mode is only learned for $p > 200$. $k=4$ mode is not learned at all in this range. Our theory again perfectly fits the experiments.

Lastly, we show that our theory also works for composite functions that contain many different degree spherical harmonics. In this setup, we randomly initialize a two layer teacher neural network and train a student neural network
\begin{equation}\label{student_teacher}
    f^*(\mathbf{x}) = \mathbf{\overline{r}}^\top \sigma(\mathbf{\overline{\Theta}} \mathbf{x}),  \quad f(\mathbf{x}) = \mathbf{{r}}^\top \sigma(\mathbf{{\Theta}} \mathbf{x}),
\end{equation}
where $\mathbf{\Theta},\mathbf{\overline{\Theta}} \in \mathbb{R}^{M \times d}$ are the feedforward weights for the student and teacher respectively, $\sigma$ is an activation function and $\mathbf{r},\mathbf{\overline{r}} \in \mathbb{R}^M$ are the student and teacher readout weights. Chosen in this way with ReLU activations, the teacher is composed of spherical harmonics of many different degrees (Section \ref{SINeuralExperiment} in SI). The total generalization error for this teacher student setup as well as the theoretical prediction of our theory is provided in Figure \ref{fig:net_c} for $d=25$, $N=8000$. They agree excellently. Results from additional neural network simulations are provided in Section \ref{SINeuralExperiment} of the SI.

\subsection{Gaussian Kernel Regression and Interpolation}

We next test our theory on another widely-used kernel. The setting where the probability measure and kernel are Gaussian, $K(\mathbf{x},\mathbf{x}') = e^{-\frac{1}{2\ell^2} ||\mathbf{x}-\mathbf{x}'||^2}$, allows analytical computation of the eigenspectrum, $\lbrace \lambda_k\rbrace$ \cite{GPMLRasmussen}. In $d$ dimensions, the $k$-th distinct eigenvalue corresponds to a set of $N(d,k) = \binom{d+k-1}{k} \sim {d^k}/{k!}$ degenerate eigenmodes. The spectrum itself decays exponentially. 

In Figure \ref{fig:kernel_a}, experimental learning curves for $d=20$ dimensional standard normal random vector data and a Gaussian kernel with $\ell = 50$ are compared to our theoretical predictions for varying ridge parameters $\lambda$. The target function $f^*(\mathbf{x})$ is constructed with the same rule we used for the NTK experiments, shown in eq. \ref{synthetic_teacher}. When $\lambda$ is small, sharp drops in the generalization error occur when $p \approx N(d,k)$ for $k=1,2$. These drops are suppressed by the explicit regularization $\lambda$.

\subsection{MNIST: Discrete Data Measure and Kernel PCA}\label{SecMNIST}

We can also test our theory for finite datasets by defining a probability measure with equal point mass on each of the data points $\{\mathbf{x}_i\}_{i=1}^{\tilde{p}}$ in the dataset (including training and test sets):
\begin{equation}
    p(\mathbf{x}) = \frac{1}{\tilde{p}} \sum_{i=1}^{\tilde{p}} \delta(\mathbf{x}-\mathbf{x}_i).
\end{equation}
With this measure, the eigenvalue problem \eqref{eq:eigenvalue} becomes a $\tilde{p} \times \tilde{p}$ kernel PCA problem (see SI \ref{SI_KPCA})
\begin{equation}
    \mathbf{K}\bm{\Phi}^{\top} = \tilde{p} \bm{\Phi}^{\top} \bm{\Lambda}.
\end{equation}
Once the eigenvalues $\mathbf{\Lambda}$ and eigenvectors $\bm{\Phi}^{\top}$ have been identified, we 
compute the target function coefficients by projecting the target data $\mathbf{y}_c$ onto these principal components $\mathbf{\overline{w}_c} = \bm{\Lambda}^{-1/2} \bm{\Phi} \mathbf{y}_c$ for each target $c=1,\ldots,C$. Once all of these ingredients are obtained, theoretical learning curves can be computed using Algorithm \ref{alg:learning_curve} and  multiple class formalism described in Section \ref{Multiple_Classes}, providing estimates of the error on the entire dataset incurred when training with a subsample of $p < \tilde{p}$ data points. An example where the discrete measure is taken as $\tilde{p} =8000$ images of handwritten digits from MNIST \cite{MNIST} and the kernel is NTK with 3 layers is provided in Figures \ref{fig:kernel_b}and \ref{fig:kernel_c}. For total generalization error, we find perfect agreement between kernel regression and neural network experiments, and our theory.

\section{Conclusion}

In this paper, we presented an approximate theory of the average generalization performance for kernel regression. We studied our theory in the ridgeless limit to explain the behavior of trained neural networks in the infinite width limit \cite{jacot2018neural,arora2019exact,lee2019wide}. We demonstrated how the RKHS eigenspectrum of NTK encodes a preferential bias to learn high spectral modes only after the sample size $p$ is sufficiently large. Our theory fits kernel regression experiments remarkably well. We further experimentally verified that the theoretical learning curves obtained in the infinite width limit provide a good approximation of the learning curves for wide but finite-width neural networks. Our MNIST result suggests that our theory can be applied to datasets with practical value.



\section*{Acknowledgements} We thank Matthieu Wyart and Stefano Spigler for comments and pointing to a recent version of their paper \cite{spigler2019asymptotic} with an independent derivation of the generalization error scaling for power law kernel and target spectra (see \eqref{scalings}). C. Pehlevan thanks the Harvard Data Science Initiative, Google and Intel for support.



\bibliography{bibliography}
\bibliographystyle{icml2020}

\newpage

\setcounter{section}{0}
\setcounter{equation}{0}
\setcounter{figure}{0}
\renewcommand{\theequation}{SI.\arabic{equation}}
\renewcommand{\thefigure}{SI.\arabic{figure}}

\section{Background on Kernel Machines}\label{SIKernelMachines}

\subsection{Reproducing Kernel Hilbert Space}

Let $\mathcal{X} \subseteq \mathbb{R}^d$ and $p(\mathbf{x})$ be a probability distribution over $\mathcal{X}$. Let $\mathcal{H}$ be a Hilbert space with inner product $\left< \cdot , \cdot \right>_{\mathcal{H}}$. A kernel $K(\cdot, \cdot)$ is said to be reproducing for $\mathcal{H}$ if function evaluation at any $\mathbf{x} \in \mathcal{X}$ is the equivalent to the Hilbert inner product with $K(\cdot, \mathbf{x})$: $K$ is reproducing for $\mathcal{H}$ if for all $g \in \mathcal{H}$ and all $\mathbf{x} \in \mathcal{X}$
\begin{equation}
    \left< K(\cdot,\mathbf{x}), g \right>_{\mathcal{H}} = g(\mathbf{x}).
\end{equation}
If such a kernel exists for a Hilbert space, then it is unique and defined as the reproducing kernel for the RKHS \cite{poggio_regnets,scholkopf_smola}. 

\subsection{Mercer's Theorem}

Let $\mathcal{H}$ be a RKHS with kernel $K$. Mercer's theorem \cite{mercer1909xvi, GPMLRasmussen} allows the eigendecomposition of $K$
\begin{equation}
    K(\mathbf{x}, \mathbf{x}') = \sum_\rho \lambda_\rho \phi_\rho(\mathbf{x}) \phi_\rho(\mathbf{x}),
\end{equation}
where the eigenvalue statement is
\begin{equation}
    \int d\mathbf{x}' p(\mathbf{x}') K(\mathbf{x},\mathbf{x}') \phi_\rho(\mathbf{x}') = \lambda_\rho \phi_\rho(\mathbf{x}).
\end{equation}

\subsection{Representer Theorem}
Let $\mathcal{H}$ be a RKHS with inner product $\left< .,.\right>_{\mathcal{H}}$. Consider the regularized learning problem
\begin{equation}
    \text{min}_{f\in \mathcal{H}} \hat{\mathcal{L}}[f] + \lambda ||f||_{\mathcal{H}}^2,
\end{equation}
where $\hat{\mathcal{L}}[f]$ is an empirical cost defined on the discrete support of the dataset and $\lambda > 0$. The optimal solution to the optimization problem above can always be written as \cite{scholkopf_smola}
\begin{equation}
    f(x) = \sum_{i=1}^p \alpha_i K(x_i,x).
\end{equation}

\subsection{Solution to Least Squares}

Specializing to the case of least squares regression, let 
\begin{equation}
    \hat{\mathcal{L}}[f] = \sum_{i=1}^p (f(\mathbf{x}_i) - y_i)^2.
\end{equation}
Using the representer theorem, we may reformulate the entire objective in terms of the $p$ coefficients $\alpha_i$ 
\begin{align}
    \mathcal{L}[f] &= \sum_{i=1}^p (f(\mathbf{x}_i) - y_i)^2 + \lambda ||f||_{\mathcal{H}}^2 \nonumber
    \\
    &= \sum_{i=1}^p (\sum_{j=1}^p \alpha_j K(\mathbf{x}_i,\mathbf{x}_j) - y_i)^2 
    \nonumber \\
    &\quad + \lambda \sum_{ij}\alpha_i \alpha_j \Big< K(\mathbf{x}_i,\cdot), K(\mathbf{x}_j,\cdot) \Big>_{\mathcal{H}}
   \nonumber \\
   & = \bm{\alpha}^\top \mathbf{K}^2 \bm{\alpha} - 2 \mathbf{y}^\top \mathbf{K} \bm{\alpha} + \mathbf{y}^\top \mathbf{y} + \lambda \bm{\alpha}^\top \mathbf{K} \bm{\alpha}.
\end{align}
Optimizing this loss with respect to $\bm{\alpha}$ gives
\begin{equation}
    \bm{\alpha} = (\mathbf{K} + \lambda \mathbf{I})^{-1} \mathbf{y}.
\end{equation}
Therefore the optimal function evaluated at a test point is
\begin{equation}
    f(\mathbf{x}) = \bm{\alpha}^\top \mathbf{k}(\mathbf{x}) = \mathbf{y}^\top (\mathbf{K} + \lambda \mathbf{I})^{-1} \mathbf{k}(\mathbf{x}).
\end{equation}

\section{Derivation of the Generalization Error}\label{SIDerivGenErr}

Let the RKHS $\mathcal{H}$ have eigenvalues $\lambda_\rho$ for $\rho \in \mathbb{Z}^+$. Define $\psi_\rho(\mathbf{x}) = \sqrt{\lambda_\rho} \phi_\rho(\mathbf{x})$, where $\phi_\rho$ are the eigenfunctions of the reproducing kernel for $\mathcal{H}$. Let the target function have the following expansion in terms of the kernel eigenfunctions $f^*(\mathbf{x}) = \sum_{\rho} \overline{w}_\rho \psi_\rho(\mathbf{x})$. Define the design matrices $\mathbf{\Phi}_{\rho, i} = \phi_\rho(\mathbf{x}_i)$ and $\mathbf{\Lambda}_{\rho \gamma} = \lambda_\rho \delta_{\rho \gamma}$.
Then the average generalization error for kernel regression is
%
\begin{equation}\label{mainEg_SI}
    E_g = {\rm Tr} \left(\mathbf{D} \left< \mathbf{G}^2 \right>_{\{\mathbf{x}_i \}} \right)
\end{equation}
where 
\begin{align}
    \mathbf{G} = \left (\frac{1}{\lambda} \mathbf{\Phi} \mathbf{\Phi}^\top + \mathbf{\Lambda}^{-1} \right)^{-1},\quad  \mathbf{\Phi} = \mathbf{\Lambda}^{-1/2} \mathbf{\Psi}.
\end{align}
 and 
 \begin{equation}
    \mathbf{D} = \mathbf{\Lambda}^{-1/2} \left< \mathbf{\overline{w}}\mathbf{\overline{w}}^\top \right>_{\mathbf{\overline{w}}} \mathbf{\Lambda}^{-1/2}.
\end{equation}
\begin{proof}
Define the student's eigenfunction expansion $f(\mathbf{x}) = \sum_\rho w_\rho \psi_\rho(\mathbf{x})$ and decompose the risk in the basis of eigenfunctions:
\begin{align}
E_g(\{\mathbf{x}_i\}, f^*)& = \left< (f( \mathbf{x} )- y(\mathbf{x}))^2 \right>_{\mathbf{x}} \nonumber
\\
&= \sum_{\rho,\gamma} (w_\rho - \overline{w}_\rho) (w_\gamma - \overline{w}_\gamma) \left< \psi_\rho(\mathbf{x}) \psi_\gamma(\mathbf{x}) \right>_{\mathbf{x}} \nonumber
\\
&= \sum_{\rho} \lambda_\rho (w_\rho - \overline{w}_\rho)^2 \nonumber
\\
&= (\mathbf{w} - \mathbf{\overline{w}})^\top \mathbf{\Lambda} (\mathbf{w} - \mathbf{\overline{w}}).
\end{align}
Next, it suffices to calculate the weights $\mathbf{w}$ learned through kernel regression. Define a matrix with elements $\mathbf{\Psi}_{\rho, i} = \psi_\rho(\mathbf{x}_i)$. The training error for kernel regression is
\begin{equation}
    E_{tr} = ||\mathbf{\Psi}^\top \mathbf{w} - \mathbf{y}||^2 + \lambda ||\mathbf{w}||_2^2
\end{equation}

The $\ell_2$ norm on $\mathbf{w}$ is equivalent to the Hilbert norm on the student function. If $f(\mathbf{x}) = \sum_\rho w_\rho \psi_\rho(\mathbf{x})$ then
\begin{align}
    \nonumber
    ||f||^2_{\mathcal{H}} &= \left< f, f \right>_{\mathcal{H}}
    \\
    &= \sum_{\rho \gamma} w_\rho w_\gamma \left< \psi_\rho(\cdot), \psi_\gamma(\cdot) \right>_{\mathcal{H}} = \sum_\rho w_\rho^2,
\end{align}
since $\left< \psi_\rho(\cdot), \psi_\gamma(\cdot) \right>_{\mathcal{H}} = \delta_{\rho,\gamma}$ \cite{bietti2019inductive}. This fact can be verified by invoking the reproducing property of the kernel and it's Mercer decomposition. Let $g(\cdot) = \sum_{\rho} a_\rho \psi_\rho(\cdot)$. By the reproducing property
\begin{align}
    \left< K(\cdot, \mathbf{x}), g(\cdot) \right>_{\mathcal{H}} &= \sum_{\rho,\gamma} a_\gamma \psi_\rho(\mathbf{x}) \left< \psi_\rho(\cdot),\psi_\gamma(\cdot) \right>_{\mathcal{H}}\nonumber
    \\
    &= g(\mathbf{x}) =  \sum_\rho a_\rho \psi_\rho(\mathbf{x})
\end{align}
Demanding equality of each term, we find
\begin{equation}
    \sum_\gamma a_\gamma \left< \psi_\rho(\cdot),\psi_\gamma(\cdot) \right>_{\mathcal{H}} = a_\rho
\end{equation}
Due to the arbitrariness of $a_\rho$, we must have $\left< \psi_\rho(\cdot), \psi_\gamma(\cdot) \right>_{\mathcal{H}} = \delta_{\rho,\gamma}$. We stress the difference between the action of the Hilbert inner product and averaging feature functions over a dataset $\left< \psi_\rho(\mathbf{x}) \psi_\gamma(\mathbf{x}) \right>_{\mathbf{x}} = \lambda_\rho \delta_{\rho,\gamma}$ which produce different results. We will always decorate angular brackets with $\mathcal{H}$ to denote Hilbert inner product.

The training error has a unique minimum
\begin{align}
    \mathbf{w} &= (\mathbf{\Psi} \mathbf{\Psi}^\top +\lambda \mathbf{I})^{-1} \mathbf{\Psi} \mathbf{y} = (\mathbf{\Psi} \mathbf{\Psi}^\top +\lambda \mathbf{I})^{-1} \mathbf{\Psi} \mathbf{\Psi}^\top \mathbf{\overline{w}}
    \nonumber \\
    &= \mathbf{\overline{w}} - \lambda (\mathbf{\Psi} \mathbf{\Psi}^\top +\lambda \mathbf{I})^{-1} \mathbf{\overline{w}},
\end{align}
where the target function is produced according to $\mathbf{y} = \mathbf{\Psi}^\top \mathbf{\overline{w}}$.

Plugging in the $\mathbf{w}$ that minimizes the training error into the formula for the generalization error, we find 
\begin{equation}\label{SI_gen_error_deriv}
    E_g(\{\mathbf{x}_i\}, \mathbf{\overline{w}}) = \lambda^2 \left< \mathbf{\overline{w}} (\mathbf{\Psi} \mathbf{\Psi}^\top +\lambda \mathbf{I})^{-1} \mathbf{\Lambda} (\mathbf{\Psi} \mathbf{\Psi}^\top +\lambda \mathbf{I})^{-1} \mathbf{\overline{w}}  \right>.
\end{equation}
Defining 
\begin{equation}
    \mathbf{G} = \lambda \mathbf{\Lambda}^{1/2} (\mathbf{\Psi} \mathbf{\Psi}^\top +\lambda \mathbf{I})^{-1} \mathbf{\Lambda}^{1/2} =  \left( \frac{1}{\lambda} \mathbf{\Phi} \mathbf{\Phi}^\top + \mathbf{\Lambda}^{-1}  \right)^{-1},
\end{equation}
and 
\begin{equation}
    \mathbf{D} = \mathbf{\Lambda}^{-1/2} \left< \mathbf{\overline{w}} \mathbf{\overline{w}}^\top \right> \mathbf{\Lambda}^{-1/2},
\end{equation}
and identifying the terms in \eqref{SI_gen_error_deriv} with these definitions, we obtain the desired result. Then each component of the mode error is given by:
\begin{equation}
    E_\rho = \sum_\gamma \mathbf{D}_{\rho,\gamma} \braket{\mathbf{G
    }^2_{\gamma,\rho}}
\end{equation}
\end{proof}

\section{Solution of the PDE Using Method of Characteristics}\label{SISolutionPDE}




Here we derive the solution to the PDE in equation \ref{pde} of the main text by adapting the method used by \cite{sollich1999learning}. We will prove both Propositions \ref{prop2} and \ref{prop3}.

Let 
\begin{align}
    g_\rho(p,v)  \equiv \left< \mathbf{\tilde{G}}(p,v)_{\rho \rho} \right>,
\end{align} 
and 
\begin{align}\label{tpv}
t(p,v) \equiv \text{Tr} \left< \mathbf{G}(p,v)\right> =  \sum_\rho g_\rho(p,v).
\end{align}
It follows from equation \ref{pde} that $t$ obeys the PDE
\begin{equation}\label{tpde}
    \frac{\partial t(p,v)}{\partial p} = \frac{1}{\lambda + t} \frac{\partial t(p,v)}{\partial v},
\end{equation}
with an initial condition $t(0,v) = \text{Tr}( \mathbf{\Lambda}^{-1} + v \mathbf{I})^{-1}$. The solution to first order PDEs of the form is given by the method of characteristics \cite{garfken67:math}, which we describe below, and prove Proposition \ref{prop2}.

\begin{proof}[Proof of Proposition 2]
The solution to \eqref{tpde} is a surface $(t,p,v) \subset \mathbb{R}^{3}$ that passes through the line $(\text{Tr}( \mathbf{\Lambda}^{-1} + v \mathbf{I})^{-1}, 0, v)$ and satisfies the PDE at all points.
The tangent plane to the solution surface at a point $(t,p,v)$ is $\text{span}\{ (\frac{\partial t}{\partial p}, 1, 0),( \frac{\partial t}{\partial v}, 0, 1) \}$. Therefore a vector $\mathbf{a} = (a_t, a_p, a_v) \in \mathbb{R}^3$ normal to the solution surface must satisfy
\begin{align}
    \nonumber
    a_t \frac{\partial t}{\partial p} + a_p = 0,\quad 
    a_t \frac{\partial t}{\partial v} + a_v = 0.
\end{align}
One such normal vector is $(-1,\frac{\partial t}{\partial p}, \frac{\partial t}{\partial v})$.

The PDE can be written as a dot product involving this normal vector,
\begin{equation}
    \left(-1,\frac{\partial t}{\partial p}, \frac{\partial t}{\partial v} \right) \cdot \left (0, 1, - \frac{1}{\lambda+t} \right) = 0,
\end{equation}
demonstrating that $(0, 1, - \frac{1}{\lambda+t} )$ is tangent to the solution surface. This allows us to parameterize one dimensional curves along the solution in these tangent directions. Such curves are known as characteristics. Introducing a parameter $s \in \mathbb{R}$ that varies along the one dimensional characteristic curves, we get
\begin{align}
    \frac{d t}{d s} = 0,\quad
    \frac{d p}{d s} = 1, \quad
    \frac{d v}{d s} = - \frac{1}{\lambda + t}.
\end{align}
The first of these equations indicate that $t$ is constant along each characteristic curve. Integrating along the parameter, $p = s + p_0$ and $v = - \frac{s}{\lambda+t} + v_0$ where $p_0$ is the value of $p$ when $s=0$ and $v_0$ is the value of $v$ at $s=0$. Without loss of generality, take $p_0 = 0$ so that $s=p$. At $s=0$, we have our initial condition
\begin{equation}
    t(0,v) = \text{Tr} \left( \mathbf{\Lambda}^{-1} + v_0 \mathbf{I} \right)^{-1}.
\end{equation}
Since $t$ takes on the same value for each characteristic
\begin{equation}\label{timplicit}
    t(p,v) = \text{Tr} \left( \mathbf{\Lambda}^{-1} + \left(v + \frac{p}{\lambda+t(p,v)} \right) \mathbf{I} \right)^{-1},
\end{equation}
which gives an implicit solution for $t(p,v)$. 
Now that we have solved for $t(p,v)$, remembering \eqref{tpv}, we may write
\begin{equation}
    g_\rho(p,v) = \left(\frac{1}{\lambda_\rho} + v + \frac{p}{\lambda + t(p,v)}\right)^{-1}.
\end{equation}
This equation proves Proposition \ref{prop2} of the main text.
\end{proof}

Next, we compute the modal generalization errors $E_{\rho}$ and prove Proposition \ref{prop3}.

\begin{proof}[Proof of Proposition 3]Computing generalization error of kernel regression requires the differentiation with respect to $v$ at $v=0$ (eq.s \eqref{mainEg} and \eqref{mainRel} of main text). Since $\left< \mathbf{G}^2 \right>$ is diagonal, the mode errors only depend on the diagonals of $\mathbf{D}$ and on $\left< \mathbf{G}_{\rho,\rho}^2 \right> = -\frac{\partial g_\rho }{\partial v}|_{v=0}$: 
\begin{align}
    E_\rho &= \sum_{\gamma} \mathbf{D}_{\rho,\gamma} \left< \mathbf{G}^2_{\gamma,\rho} \right> =  - \frac{\braket{\overline{w}_\rho^2}}{\lambda_\rho}\left. \frac{\partial g_\rho}{\partial v}\right|_{v=0}.
\end{align}

We proceed with calculating the derivative in the above equation.
\begin{align}
    \frac{\partial g_\rho(p,0) }{\partial v} &= -\left(\frac{1}{\lambda_\rho} + \frac{p}{\lambda + t(p,0)} \right)^{-2}  \nonumber \\
    &\quad\times \left( 1 - \frac{p}{(\lambda + t)^2} \frac{\partial t(p,0)}{\partial v} \right).
\end{align}
We need to calculate $\frac{\partial t(p,v)}{\partial v}|_{v=0}$
\begin{equation}
    \frac{\partial t(p,0)}{\partial v} = - 
    \gamma \left( 1 - \frac{p}{(\lambda + t)^2} \frac{\partial t(p,0)}{\partial v} \right),
\end{equation}
where 
\begin{align}\label{gammadef}
\gamma \equiv  \sum_\rho \left(\frac{1}{\lambda_\rho} + \frac{p}{\lambda + t(p,0)} \right)^{-2}.
\end{align}
Solving for the derivative, we get
\begin{equation}
    \frac{\partial t(p,0)}{\partial v} = - \frac{\gamma}{1 - \gamma \frac{p}{(\lambda+t)^2}}, 
\end{equation}
and
\begin{equation}
    \frac{\partial g_\rho(p,0) }{\partial v} = -\left(\frac{1}{\lambda_\rho} + \frac{p}{\lambda + t} \right)^{-2} \left(1 - \frac{\gamma p}{(\lambda+t)^2} \right)^{-1}.
\end{equation}
The error in mode $\rho$ is therefore
\begin{align}\label{mode_err}
    E_\rho =  \frac{\braket{\overline{w}_\rho^2}}{\lambda_\rho}\left(\frac{1}{\lambda_\rho} + \frac{p}{\lambda+t(p)}\right)^{-2} \left(1 - \frac{p \gamma(p)}{(\lambda + t(p))^2}\right)^{-1},
\end{align}
so it suffices to numerically solve for $t(p,0)$ to recover predictions of the mode errors. Equations \eqref{timplicit} (evaluated at $v=0$), \eqref{gammadef} and \eqref{mode_err} collectively prove Proposition \ref{prop3}.
\end{proof}

\section{Learning Curve for Power Law Spectra}\label{SIPowerLaw}

For $\lambda > 0$, the mode errors asymptotically satisfy $E_\rho \sim \mathcal{O}(p^{-2})$ since $\frac{p}{\lambda+t} \sim \frac{p}{\lambda}$ and $\frac{(\lambda+t)^2}{(\lambda+t)^2-\gamma p} \sim \mathcal{O}_p(1)$ (see below). Although each mode error decays asymptotically like $p^{-2}$, the total generalization error can have nontrivial scaling with $p$ that depends on both the kernel and the target function.

To illustrate the dependence of the learning curves on the choice of kernel and target function, we consider a case where both have power law spectra. Specifically, we assume that $\lambda_\rho = \rho^{-b}$ and $a_\rho^2 \equiv \overline{w}_\rho^2 \lambda_\rho = \rho^{-a}$ for $\rho = 1,2,...$. We introduce the variable $z=t+\lambda$ to simplify the computations below. We further approximate the sums over modes with integrals
\begin{align}
    E_g &\approx \frac{z^2}{z^2- p \gamma}  \int_1^\infty \frac{d\rho \ \rho^{-a}}{\left(\frac{p}{z}  \rho^{-b} + 1 \right)^2}. 
\end{align}
We use the same approximation technique to study the behavior of $z(p)$
\begin{align}
    z &= \lambda + \frac{z}{p} \int_1^\infty \frac{d\rho}{1+\frac{z}{p} \rho^b} \nonumber
   = \lambda + \left(\frac{z}{p} \right)^{1-\frac{1}{b}} \int_{(z/p)^{1/b}}^\infty \frac{du}{1+u^b} \\
    & = \lambda + \left(\frac{z}{p} \right)^{1-\frac{1}{b}} F(b, p,z),
\end{align}
where $F(b,p,z) = \int_{(z/p)^{1/b}}^\infty \frac{du}{1+u^b}$. If $p \gg \lambda^{-1/(b-1)}$ then $z \approx \lambda$, otherwise $z \approx p^{1-b} F(b,p,z)^b$. Further, the scaling $z \sim \mathcal{O}(p^{1-b})$ is self-consistent since the lower endpoint of integration $(z/p)^{1/b} \sim p^{-1} \to 0$ so $F(b,z,p)$ approaches a constant $F(b)$ for $p \to \infty$
\begin{equation}
    z \sim p^{1-b} F(b)^b \ , \ F(b,z,p) \sim F(b) = \int_0^\infty \frac{du}{1+u^b}.
\end{equation}

We similarly find that $p \gamma(p) \sim \mathcal{O}(p^{2-2b})$ if $p \ll \lambda^{-1/(b-1)}$. The mode-independent prefactor is approximately constant $\frac{z^2}{z^2-\gamma p} \sim \mathcal{O}_p(1)$. 

We can use all of these facts to identify scalings of $E_g$. We will first consider the case where $p \ll \lambda^{-1/(b-1)}$:
\begin{align}
    E_g &\sim \int_{1}^{\infty} \frac{d\rho \rho^{-a} }{(p^b \rho^{-b} + 1)^2}  \nonumber
    \\
    &\approx p^{-2b} \int_1^{p} d\rho \rho^{-a+2b} + \int_p^\infty d\rho \rho^{-a} \nonumber
    \\
    & = \frac{1}{a-1-2b} p^{-2b} + \frac{2b}{(a-1)(2b+1-a)} p^{-(a-1)}.
\end{align}
If $2b > a-1$ then the second term dominates, indicating that higher frequency modes $k > p$ provide a greater contribution to the error due to the slow decay in the target power. In this case $E_g \sim p^{-(a-1)} $. If, on the other hand, $2b < a-1$ then lower frequency modes $k < p$ dominate the error and $E_g \sim p^{-2b}$. 

Now, suppose that $p > \lambda^{-1/(b-1)}$.  In this regime 
\begin{align}
    E_g &\sim \int_1^\infty \frac{d\rho \rho^{-a}}{( \frac{p}{\lambda} \rho^{-b} +1 )^2} \nonumber
    \\
    &\approx \frac{\lambda^2}{p^2} \int_1^{(p/\lambda)^{1/b}} d\rho \ \rho^{2b-a} + \int_{(p/\lambda)^{1/b}}^\infty d\rho \rho^{-a} \nonumber
    \\
    &= \frac{\lambda^2}{p^2} \frac{1}{2b-a+1} \left[ \left(\frac{p}{\lambda} \right)^{(2b-a+1)/b} - 1 \right] \nonumber
    \\
    &+ \frac{1}{a-1} \left( \frac{p}{\lambda} \right)^{(1-a)/b}.
\end{align}
Here there are two possible scalings. If $2b > a-1$ then $E_g \sim p^{-(a-1)/b}$ while $2b < a-1$ implies $E_g \sim p^{-2}$. 

So the total error scales like
\begin{align}\label{scalings}
    &E_g \sim p^{- \min\{a-1, 2b\}} \ , \ &p < \lambda^{-1/(b-1)} \nonumber
    \\
   & E_g \sim  p^{- \min\{a-1,2b \}/b} \ , \ &p > \lambda^{-1/(b-1)}.
\end{align}

\begin{figure*}[t]
\subfigure[$z = t+\lambda$]{\label{sfig:power_law_scalings}\includegraphics[width=0.5\linewidth]{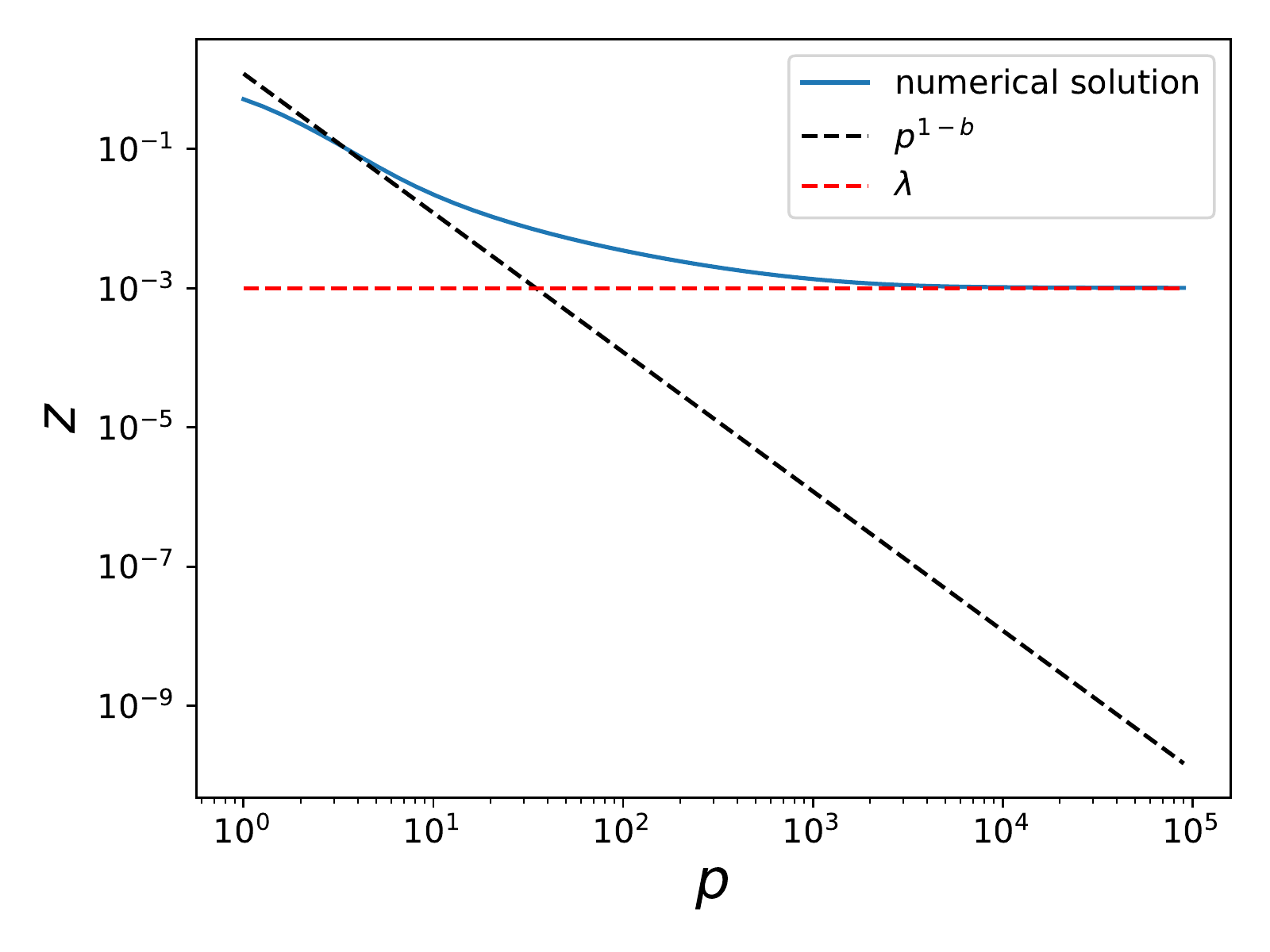}}
\subfigure[$E_g(p)$]{\label{sfig:power_law_scalings2}\includegraphics[width=0.5\linewidth]{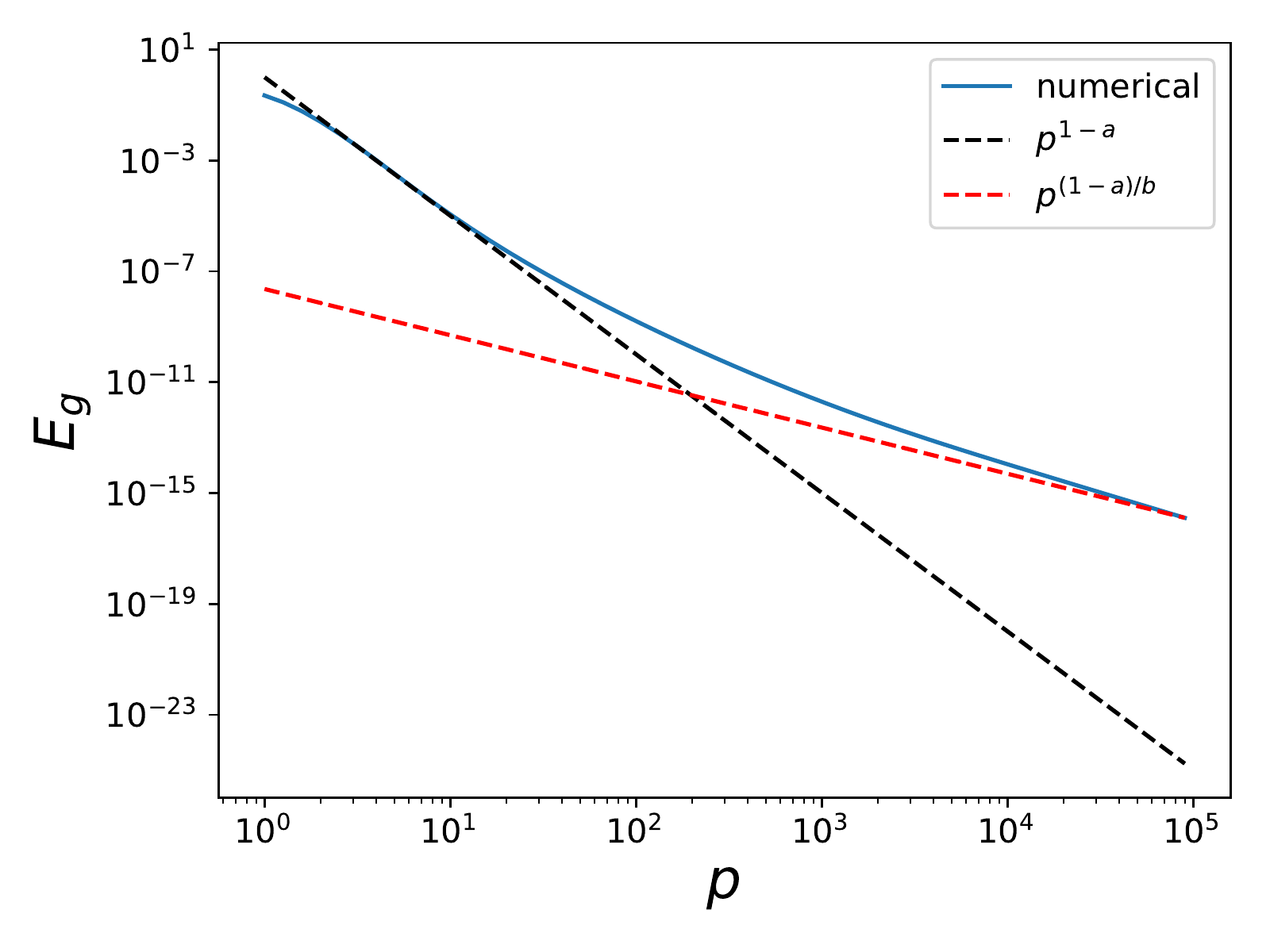}}
\caption{Approximate scaling of learning curve for spectra that decay as power laws $\lambda_k \sim k^{-b}$ and $a_k^2 \equiv \overline{w}_k^2 \lambda_k = k^{-a}$. Figure (a) shows a comparison of the numerical solution to the implicit equation for $t+\lambda$ as a function of $p$ and its comparison to approximate scalings. There are two regimes which are separated by $p \approx \lambda^{- 1/(b-1)}$. For small $p$, $z \sim p^{1-b}$ but for large $p$, $z \sim \lambda$. The total generalization error is shown in (b) which scales like $p^{1-a}$ for small $p$ and $p^{(1-a)/b}$ for large p. }
\label{fig:power_law_scaling}
\end{figure*}

A verification of this scaling is provided in Figure \ref{fig:power_law_scaling}, which shows the behavior of $z$ and $E_g$ in these two regimes. When the explicit regularization is low (or zero) ($p<\lambda^{-1/(b-1)}$), our equations reproduce the power law scalings derived with Fourier analysis in \cite{spigler2019asymptotic}\footnote{We note that in a recent version of their paper, \citet{spigler2019} used our formalism to independently derive the scalings in \eqref{scalings} for the ridgeless ($\lambda=0$) case. Our calculation in an earlier preprint had missed the possible $\sim p^{-2b}$ and $\sim p^{-2}$ scalings, which we corrected after their paper.}.

The slower asymptotic decays in generalization error when explicit regularization $\lambda$ is large relative to the sample size indicates that explicit regularization hurts performance. The decay exponents also indicate that the RKHS eigenspectrum should decay with exponent at least as large as $b^* > \frac{a-1}{2}$ for optimal asymptotics. Kernels with slow decays in their RKHS spectra induce larger errors.

\section{Replica Calculation}\label{SIReplicaCalc}

In this section, we present the replica trick and the saddle-point approximation summarized in main text Section \ref{sec:main_replica}. Our goal is to show that the continuous approximation of the main paper and previous section can be interpreted as a finite size saddle-point approximation to the replicated system under a replica symmetry ansatz. We will present a detailed treatment of the thermodynamic limit and the replica symmetric ansatz in a different paper.

Let $\mathbf{\tilde{G}}(p,v) = \left(\frac{1}{\lambda} \mathbf{\Phi} \mathbf{\Phi}^\top + \mathbf{\Lambda}^{-1} + v \mathbf{I} \right)^{-1}$. 
To obtain the average elements $\left< \mathbf{\tilde{G}}(p,v)_{\rho,\gamma} \right>$ we will use a Gaussian integral representation of the matrix inverse
\begin{align}
    &\left< \mathbf{\tilde{G}}(p,v)_{\rho,\gamma} \right> \nonumber\\
    &= \frac{\partial^2}{\partial h_\rho \partial h_\gamma} \left< \frac{1}{Z} \int \ d\mathbf{u}  e^{-\frac{1}{2} \mathbf{u}(\frac{1}{\lambda} \mathbf{\Phi}\mathbf{\Phi}^\top + \mathbf{\Lambda}^{-1} +v\mathbf{I}) \mathbf{u} + \mathbf{h} \cdot \mathbf{u}} \right>_{\mathbf{\Phi}},
\end{align}
where 
\begin{equation}
    Z = \int d\mathbf{u} \ e^{-\frac{1}{2} \mathbf{u}(\frac{1}{\lambda} \mathbf{\Phi}\mathbf{\Phi}^\top + \mathbf{\Lambda}^{-1} +v\mathbf{I}) \mathbf{u} },
\end{equation}
and make use of the identity $Z^{-1} = \lim_{n \to 0} Z^{n-1}$ to rewrite the entire average in the form
\begin{align}
    R(\mathbf{h}) = \int \prod_{a=1}^n d\mathbf{u}^a \left< e^{-\frac{1}{2} \sum_a \mathbf{u}^a(\frac{1}{\lambda} \mathbf{\Phi}\mathbf{\Phi}^\top + \mathbf{\Lambda}^{-1} +v\mathbf{I}) \mathbf{u}^a + \mathbf{h} \cdot \mathbf{u}^{(1)}} \right>
\end{align}
with the identification that 
\begin{equation}
    \left< \mathbf{\tilde{G}}(p,v)_{\rho,\gamma} \right> = \frac{\partial^2}{\partial h_\rho \partial h_\gamma}\lim_{n \to 0} R(\mathbf{h})|_{\mathbf{h}=0}.
\end{equation}
Following the replica method from the physics of disordered systems, we will first restrict ourselves to integer $n$ and then analytically continue the resulting expressions to take the limit of $n \to 0$. 

Averaging over the quenched disorder (dataset) with the assumption that the residual error $(\mathbf{w} - \mathbf{\overline{w}}) \cdot \mathbf{\Psi}(\mathbf{x}_i)$ is a Gaussian process, we find
\begin{equation}
    \left< e^{-\frac{1}{2 \lambda} \sum_a \mathbf{u}^a \mathbf{\Phi} \mathbf{\Phi}^\top \mathbf{u}^a } \right> = e^{- \frac{p}{2} \log \det(\mathbf{I}+\frac{1}{\lambda}\mathbf{Q})},
\end{equation}
where order parameters $Q_{ab} = \mathbf{u}^a \cdot \mathbf{u}^b$ have been introduced.

To enforce the definition of these order parameters, Dirac delta functions are inserted into the expression for $R$. We then represent each delta function as a Fourier integral so that integrals over $\mathbf{u}^a$ can be computed 
\begin{equation}
    \delta( Q_{ab} - \mathbf{u}^a \cdot \mathbf{u}^b) = \int d\hat{Q}_{ab} e^{i Q_{ab} \hat{Q}_{ab} - i \hat{Q}_{ab} \mathbf{u}^a \cdot \mathbf{u}^b}.
\end{equation}
After inserting delta functions to enforce order parameter definitions, we are left with integrals over the thermal degrees of freedom
\begin{align}
    \int \prod_{a=1}^n d\mathbf{u}^a e^{ -\frac{1}{2} \sum_{a} \mathbf{u}^a\mathbf{\Lambda}^{-1} \mathbf{u}^a - i\sum_{ab}\hat{Q}_{ab} \mathbf{u}^a \mathbf{u}^b + \mathbf{u}^{(1)} \mathbf{h} } \nonumber
    \\
    = e^{- \frac{1}{2} \sum_\rho \log \det(\frac{1}{\lambda_\rho} \mathbf{I} + 2i\mathbf{\hat{Q}}) + \frac{1}{2} \sum_\rho h_\rho^2 (\frac{1}{\lambda_\rho} \mathbf{I} + 2i\mathbf{\hat{Q}})^{-1}_{11}}.
\end{align}

We now make a replica symmetric ansatz $Q_{ab} = q \delta_{ab} + q_0$ and $2 i \hat{Q}_{ab} = \hat{q} \delta_{ab} + \hat{q}_0$. Under this ansatz $R(\mathbf{h})$ can be rewritten as 
\begin{align}
    &R(\mathbf{h}) = \nonumber \\
    &\,\int d q d\hat{q} d d\hat{q} d\hat{q}_0 e^{- p n \mathcal{F}(q,q_0,\hat{q}, \hat{q}_0)} e^{\frac{1}{2} \sum_\rho h_\rho^2  (\frac{1}{\lambda_\rho}\mathbf{I} + 2i\mathbf{\hat{Q}})_{11}^{-1}},
\end{align}
where the free energy is
\begin{align}
    2 p \mathcal{F}(q,q_0,\hat{q}, \hat{q}_0) = & p \log\left(1 + \frac{q}{\lambda} \right) + p\frac{q_0}{\lambda + q} + v (q+ q_0) \nonumber
    \\
    &- (q +q_0)(\hat{q} + \hat{q}_0) + q_0 \hat{q}_0 \nonumber
    \\
    &+ \sum_{\rho}\left[ \log\left(\frac{1}{\lambda_\rho} + \hat{q}\right) + \frac{\hat{q}_0}{\frac{1}{\lambda_\rho} + \hat{q}} \right].
\end{align}

In the limit $p \to \infty$, $R(\mathbf{h})$ is dominated by the saddle point of the free energy where $\nabla \mathcal{F}(q,\hat{q},q_0,\hat{q}_0) = 0$. The saddle point equations are
\begin{align}
    \hat{q}^* &= \frac{p}{q^*+\lambda} + v, \nonumber
    \\
    q^* &= \sum_\rho \frac{1}{\frac{1}{\lambda_\rho}+\hat{q}^*} = \sum_\rho \frac{1}{\frac{1}{\lambda_\rho} + v + \frac{p}{q^*+\lambda}},\nonumber
    \\
    q_0^* &= \hat{q}_0^* = 0.
\end{align}
We see that $q^*$ is exactly equivalent to $t(p,v)$ defined in \ref{timplicit} for the continuous approximation. Under the saddle point approximation we find
\begin{equation}
    R(\mathbf{h}) \approx e^{-n p \mathcal{F}(q^*,q_0^*,\hat{q}^*, \hat{q}_0^*)} e^{\frac{1}{2} \sum_\rho h_\rho^2\frac{1}{\frac{1}{\lambda_\rho} + \hat{q}^*}}. 
\end{equation}

Taking the $n \to 0$ limit as promised, we obtain the normalized average 
\begin{equation}
    \tilde{R}(\mathbf{h}) \equiv \lim_{n \to 0} R(\mathbf{h}) = e^{\frac{1}{2} \sum_\rho h_\rho^2\frac{1}{\frac{1}{\lambda_\rho} + \hat{q}^*}},
\end{equation}
so that the matrix elements are
\begin{align}
    \left< \mathbf{\tilde{G}}(p,v)_{\rho,\gamma}\right> &= \frac{\partial^2}{\partial h_\rho \partial h_\gamma} \tilde{R}(\mathbf{h})|_{\mathbf{h}=0}= \frac{\delta_{\rho,\gamma}}{\frac{1}{\lambda_\rho} + v+ \frac{p}{\lambda+q^*}},\nonumber
    \\
    q^* &= \sum_\rho \frac{1}{\frac{1}{\lambda_\rho}+v + \frac{p}{\lambda+q^*}}.
\end{align}

Using our formula for the mode errors, we find
\begin{align}
\nonumber
    E_\rho &= \sum_{\gamma} \mathbf{D}_{\rho,\gamma} \left< \mathbf{\tilde{G}}(p,v)_{\gamma,\rho}^2 \right> 
    \\
    \nonumber
    &= - \mathbf{D}_{\rho,\rho} \frac{\partial}{\partial v} \left< \mathbf{\tilde{G}}(p,v)_{\rho,\rho} \right>|_{v=0}
    \\
    &= \frac{\left< \overline{w}_\rho^2 \right>}{\lambda_\rho} \frac{ (\lambda+q^*)^2}{(\lambda+q^*)^2-\gamma p}  \left( \frac{1}{\lambda_\rho} + \frac{p}{\lambda+q^*} \right)^{-2},
\end{align}
consistent with our result from the continuous approximation.

\section{Spectral Dependence of Learning Curves}\label{SISpectralDependency}

We want to calculate how different mode errors change as we add one more sample. We study:
\begin{equation}
    \frac{1}{2}\frac{d}{dp}\log{\frac{E_\rho}{E_\gamma}},
\end{equation}
where $E_\rho$ is given by eq. \eqref{modeError}. Evaluating the derivative, we find:
\begin{align}
    &\frac{1}{2} \frac{d}{dp}\log\left( \frac{E_\rho}{E_\gamma} \right)\nonumber\\
    &= -\left( \frac{1}{\frac{1}{\lambda_\rho} + \frac{p}{\lambda+t}} -\frac{1}{\frac{1}{\lambda_\gamma} + \frac{p}{\lambda+t}}  \right) \frac{\partial }{\partial p}\left(\frac{p}{\lambda+t}\right).
\end{align}
Using eq. \eqref{t_func},
\begin{align}\label{dtdp}
    \frac{\partial t}{\partial p} &= -\frac{\partial }{\partial p}\left(\frac{p}{\lambda+t}\right)\sum_\rho\left(\frac{1}{\lambda_\gamma} + \frac{p}{\lambda+t}\right)^{-2} \nonumber\\
    & = -\gamma \frac{\partial }{\partial p}\left(\frac{p}{\lambda+t}\right),
\end{align}
where we identified the sum with $\gamma$. Inserting this, we obtain:
\begin{align}
    &\frac{1}{2} \frac{d}{dp}\log\left( \frac{E_\rho}{E_\gamma} \right) = \left[ \frac{1}{\frac{1}{\lambda_\rho} + \frac{p}{\lambda+t}} -\frac{1}{\frac{1}{\lambda_\gamma} + \frac{p}{\lambda+t}}  \right]\frac{1}{\gamma}\frac{\partial t}{\partial p}.
\end{align}
Finally, solving for $\partial t/\partial p$ from \eqref{dtdp}, we get:
\begin{equation}
    \frac{\partial t}{\partial p} = -\frac{1}{\lambda+t}\frac{(\lambda+t)^2\gamma}{(\lambda+t)^2-p\gamma} = -\frac{1}{\lambda+t}\text{Tr}\big({\mathbf{G}^2}\big),
\end{equation}
proving that $\partial t/\partial p < 0$. Taking $\lambda_\gamma>\lambda_\rho$ without loss of generality, it follows that
\begin{equation}
    \frac{d}{dp}\log\left( \frac{E_\rho}{E_\gamma} \right) > 0\;\Rightarrow\; \frac{d}{dp} \log E_\rho > \frac{d}{dp} \log E_\gamma.
\end{equation}

\section{Spherical Harmonics}\label{SISphericalHarmonics}

Let $-\Delta$ represent the Laplace-Beltrami operator in $\mathbb{R}^d$. Spherical harmonics $\{Y_{km}\}$ in dimension $d$ are harmonic ($-\Delta Y_{km}(\mathbf{x}) = 0$), homogeneous ($Y_{km}(t\mathbf{x}) = t^{k} Y_{km}(\mathbf{x})$) polynomials that are orthonormal with respect to the uniform measure on $\mathbb{S}^{d-1}$ \cite{costasspherical,Dai_2013}. The number of spherical harmonics of degree k in dimension $d$ denoted by $N(d,k)$ is
\begin{equation}\label{eq:degens}
 N(d,k) = \frac{2k+d-2}{k}\binom{k+d-3}{k-1}.
\end{equation}

The Laplace Beltrami Operator can be decomposed into the radial and angular parts, allowing
\begin{equation}
    -\Delta = - \Delta_r - \Delta_{\mathbb{S}^{d-1}}
\end{equation}

Using this decomposition, the spherical harmonics are eigenfunctions of the surface Laplacian
\begin{equation}
    -\Delta_{\mathbb{S}^{d-1}} Y_{km}(\mathbf{x}) = k(k+d-2)Y_{km}(\mathbf{x}).
\end{equation}

The spherical harmonics are related to the Gegenbauer polynomials $\{Q_k\}$, which are orthogonal with respect to the measure $d\tau(z) = (1-z^2)^{(d-3)/2} dz$ of inner products $z = \mathbf{x}^\top \mathbf{x}'$ of uniformly sampled pairs on the sphere $\mathbf{x}, \mathbf{x}' \sim \mathbb{S}^{d-1}$. The Gegenbauer polynomials can be constructed with the Gram-Schmidt procedure and have the following properties 
\begin{align}
    Q_k&(\mathbf{x}^\top \mathbf{x}') = \frac{1}{N(d,k)} \sum_{m=1}^{N(d,k)} Y_{km}(\mathbf{x}) Y_{km}(\mathbf{x}'), \nonumber 
    \\
    &\int_{-1}^1 Q_{k}(z) Q_{\ell}(z) d\tau(z) = \frac{\omega_{d-1}}{\omega_{d-2}} \frac{\delta_{k,\ell}}{N(d,k)},
\end{align}
where $\omega_{d-1} = \frac{\pi^{d/2}}{\Gamma(d/2)}$ is the surface area of $\mathbb{S}^{d-1}$.


\section{Decomposition of Dot Product Kernels on $\mathbb{S}^{d-1}$}\label{SIDotProductKernels}

For inputs sampled from the uniform measure on $\mathbb{S}^{d-1}$, dot product kernels can be decomposed into Gegenbauer polynomials introduced in SI Section \ref{SISphericalHarmonics}. 

Let $K(\mathbf{x},\mathbf{x}') = \kappa(\mathbf{x}^\top \mathbf{x}')$. The kernel's orthogonal decomposition is 
\begin{align}
    \kappa(z) &= \sum_{k=0}^\infty \lambda_k N(d,k) Q_{k}(z), \nonumber
    \\
    \lambda_k &= \frac{\omega_{d-2}}{\omega_{d-1}} \int_{-1}^1 \kappa(z) Q_{k}(z) d\tau(z). 
\end{align}

To numerically calculate the kernel eigenvalues of $\kappa$, we use Gauss-Gegenbauer quadrature \cite{abramowitz_stegun} for the measure $d\tau(z)$ so that for a quadrature scheme of order $r$
\begin{equation}
    \int_{-1}^1 \kappa(z) Q_{k}(z) d\tau(z) \approx \sum_{i=1}^r w_i Q_{k}(z_i) \kappa(z_i),
\end{equation}
where $z_i$ are the $r$ roots of $Q_r(z)$ and the weights $w_i$ are chosen with 
\begin{equation}
    w_i = \frac{\Gamma(r+\alpha +1)^2}{\Gamma(r+2\alpha+1)} \frac{2^{2r+2\alpha+1}r!}{V_r'(z_i) V_{r+1}(z_i)},
\end{equation}
where 
\begin{equation}
    V_r(z) = 2^r r! (-1)^r Q_r(z)
\end{equation}
For our calculations we take $r=1000$.

\section{Frequency Dependence of Learning Curves in $d\to\infty$ Limit}\label{SIFrequencyDependence}

Here, we consider an informative limit where the number of input data dimension, $d$, goes to infinity.

Denoting the index $\rho = (k,m)$, we can write mode error \eqref{mode_err}, after some rearranging, as:
\begin{equation}\label{mode_err_1}
    E_{km} = \frac{(\lambda + t)^2}{1 - \frac{p \gamma}{(\lambda + t)^2}}\frac{\lambda_k\braket{\overline{w}_{km}^2}}{(\lambda+t+p\lambda_k)^2}, 
\end{equation}
where $t$ and $\gamma$, after performing the sum over degenerate indices, are:
\begin{align}
    t &= \sum_m \frac{N(d,m)(\lambda+t)\lambda_m}{\lambda+t+p\lambda_m}, \nonumber \\
    \gamma &= \sum_m \frac{N(d,m)(\lambda+t)^2\lambda_m^2}{(\lambda+t+p\lambda_m)^2}.
\end{align}
In the limit $d\to\infty$, the degeneracy factor \eqref{eq:degens} approaches to $N(d,k)  \sim \mathcal{O}(d^k)$. We note that for dot-product kernels $\lambda_k$ scales with $d$ as $\lambda_k \sim d^{-k}$ \cite{smola2001dotproduct} (Figure 1), which leads us to define the $\mathcal{O}(1)$ parameter $\bar\lambda_k = d^k\lambda_k$. Plugging these in, we get:
\begin{align}
E_{km}(g_k) &= \frac{d^{-k}(t+\lambda)^2}{1-\tilde\gamma}\frac{\bar\lambda_k\braket{\bar w_{km}^2}}{\left(t+\lambda+g_k \bar\lambda_k\right)^2}\nonumber\\
t &= \sum_m \frac{(t+\lambda)\bar\lambda_m}{t+\lambda+g_m\bar\lambda_m},\nonumber \\
\tilde\gamma &= \sum_m \frac{g_m \bar\lambda_m^2}{\left(t+\lambda+g_m\bar\lambda_m\right)^2},
\end{align}
where $g_k = p/d^k$ is the ratio of sample size to the degeneracy. Furthermore, we want to calculate the ratio $E_{km}(p)/E_{km}(0)$ to probe how much the mode errors move from their initial value:
\begin{equation}\label{mode_error_approx}
    \frac{E_{km}(p)}{E_{km}(0)} =  \frac{1}{1-\tilde\gamma}\frac{1}{\left(1+\frac{g_k \bar\lambda_k}{t+\lambda}\right)^2}
\end{equation}
Let us consider an integer $l$ such that the scaling $P = \alpha d^l$ holds. This leads to three different asymptotic behavior of $g_k$s:
\begin{align}\label{gk_scaling}
g_{k} &\sim \mathcal{O}(d^{l-k}) \gg \mathcal{O}(1),& k<l \nonumber \\
g_k &= \alpha \sim \mathcal{O}(1) ,& k=l \nonumber \\
g_{k} &\sim \mathcal{O}(d^{l-k}) \ll \mathcal{O}(1),& k>l
\end{align}

If we assume $t\sim\mathcal{O}(1)$, we get an asymptotically consistent set of equations: 
\begin{align}\label{approx_t_gamma}
t&\approx \sum_{m > l}\bar\lambda_m + a(\alpha,t,\lambda,\bar\lambda_l) \sim \mathcal{O}(1),\nonumber \\
\tilde\gamma &\approx b(\alpha,t,\lambda,\bar\lambda_l) \sim \mathcal{O}(1),
\end{align}
where $a$ and $b$ are the $l^\text{th}$ terms in the sums in $t$ and $\tilde\gamma$, respectively, and are given by:
\begin{align}
    a(\alpha,t,\lambda,\bar\lambda_l) &= \frac{(t+\lambda)\bar\lambda_l}{t+\lambda+\alpha\bar\lambda_l},\nonumber\\
    b(\alpha,t,\lambda,\bar\lambda_l) &= \frac{\alpha \bar\lambda_l^2}{\left(t+\lambda+\alpha\bar\lambda_l\right)^2}
\end{align}
Then using \eqref{mode_error_approx}, \eqref{gk_scaling} and \eqref{approx_t_gamma}, we find the errors associated to different modes as:
%
%
\begin{alignat}{2}
 &k < l,\quad &&\frac{E_{km}(\alpha)}{E_{km}(0)} \sim \mathcal{O}(d^{2(k-l)})\approx 0, \nonumber\\
 &k > l,\quad &&\frac{E_{km}(\alpha)}{E_{km}(0)} \approx \frac{1}{1-\tilde\gamma(\alpha)},\nonumber\\
 &k = l,\quad &&\frac{E_{km}(\alpha)}{E_{km}(0)} = s(\alpha)\sim\mathcal{O}(1),
 \end{alignat}
 where $s(\alpha)$ is given by:
 \begin{equation}
     s(\alpha) = \frac{1}{1-\tilde\gamma(\alpha)}\frac{1}{\left(1+\alpha\frac{ \bar\lambda_l}{t+\lambda}\right)^2}.
\end{equation}
 Note that $\lim_{\alpha\rightarrow 0}\tilde\gamma(\alpha) = \lim_{\alpha\rightarrow \infty}\tilde\gamma(\alpha) = 0$ and non-zero in between. Then, for large $\alpha$, in the limit we are considering
 \begin{alignat}{2}
 &k < l,\quad &&\frac{E_{km}(\alpha)}{E_{km}(0)} \approx 0, \nonumber\\
 &k > l,\quad &&\frac{E_{km}(\alpha)}{E_{km}(0)} \approx 1,\nonumber\\
 &k = l,\quad &&\frac{E_{km}(\alpha)}{E_{km}(0)} \approx \frac{(\lambda+\sum_{m > l}\bar\lambda_m)^2}{\bar\lambda_l^2}\frac 1{\alpha^2}.
 \end{alignat}
 




\section{Neural Tangent Kernel}\label{SINeuralTangent}

The neural tangent kernel is
\begin{equation}
    K_{\text{NTK}}(\mathbf{x},\mathbf{x}') = \sum_{i} \Big< \frac{\partial f_\theta(\mathbf{x})}{\partial \theta_i} \frac{\partial f_\theta(\mathbf{x}')}{\partial \theta_i} \Big>_{\theta}.
\end{equation}
For a neural network, it is convenient to compute this recursively in terms of the Neural Network Gaussian Process (NNGP) kernel which corresponds to only training the readout weights from the final layer \cite{jacot2018neural, arora2019exact}. We will restrict our attention to networks with zero bias and nonlinear activation function $\sigma$. Then
\begin{align}\label{eq.NTKRecursion}
&K_{NTK}^{(1)}(\mathbf{x},\mathbf{x}')\nonumber \\
&\qquad=K_{NNGP}^{(1)}(\mathbf{x},\mathbf{x}') \nonumber\\
&K_{NTK}^{(2)}(\mathbf{x},\mathbf{x}')\nonumber\\&\qquad=K_{NNGP}^{(2)}(\mathbf{x},\mathbf{x}')+K_{NTK}^{(1)}(\mathbf{x},\mathbf{x}')\dot{K}^{(2)}(\mathbf{x},\mathbf{x}') \nonumber\\ &\qquad \ldots \nonumber\\
&K_{NTK}^{(L)}(\mathbf{x},\mathbf{x}') \nonumber\\ &\qquad=  K_{NNGP}^{(L)}(\mathbf{x},\mathbf{x}')+ K_{NTK}^{(L-1)}(\mathbf{x},\mathbf{x}')\dot{K}^{(L)}(\mathbf{x},\mathbf{x}'),
\end{align}
where 
\begin{align}
\nonumber
    &K_{NNGP}^{(L)}(\mathbf{x},\mathbf{x}') = \mathbb{E}_{(\alpha,\beta) \sim p^{(L-1)}_{\mathbf{x},\mathbf{x'}} } \sigma(\alpha) \sigma(\beta),
    \\
    \nonumber
    &\dot{K}^{(L)}(\mathbf{x},\mathbf{x}') = \mathbb{E}_{(\alpha,\beta) \sim p^{(L-1)}_{\mathbf{x},\mathbf{x'}} } \dot{\sigma}(\alpha) \dot{\sigma}(\beta),
    \\
    \nonumber
    &p^{(L-1)}_{\mathbf{x},\mathbf{x'}} = \mathcal{N}\Bigg(
    \begin{pmatrix}
    0
    \\
    0
    \end{pmatrix}
    , \begin{pmatrix} K^{(L-1)}(\mathbf{x},\mathbf{x}) & K^{(L-1)}(\mathbf{x},\mathbf{x}')
    \\
    K^{(L-1)}(\mathbf{x},\mathbf{x}') & K^{(L-1)}(\mathbf{x}',\mathbf{x}')
    \end{pmatrix}
     \Bigg),
    \\
    &K^{(1)}_{NNGP}(\mathbf{x},\mathbf{x}') = \mathbf{x}^\top \mathbf{x}'.
\end{align}

If $\sigma$ is chosen to be the ReLU activation, then we can analytically simplify the expression. Defining the following function
\begin{equation}
    f(z) = \arccos\left( \frac{1}{\pi} \sqrt{1-z^2} + \left(1-\frac{1}{\pi} \arccos(z) \right) z \right),
\end{equation}
we obtain
\begin{align}
\nonumber
    K_{NNGP}^{(L)}(\mathbf{x},\mathbf{x}') &= \cos\left( f^{\circ (L-1)}(\mathbf{x}^\top \mathbf{x}') \right)
    \\
    \dot{K}_{L}(\mathbf{x},\mathbf{x}') &= \left(1-\frac{1}{\pi} f^{\circ (L-2)}(\mathbf{x}^\top \mathbf{x}') \right),
\end{align}
where $f^{\circ (L-1)}(z)$ is the function $f$ composed into itself $L-1$ times. 

This simplification gives an exact recursive formula to compute the kernel as a function of $z = \mathbf{x}^\top \mathbf{x}'$, which is what we use to compute the eigenspectrum with the quadrature scheme described in the previous section.

\section{Spectra of Fully Connected ReLU NTK}\label{SISpectraNTK}

\begin{figure}
    \centering
    \includegraphics[width=0.8\linewidth]{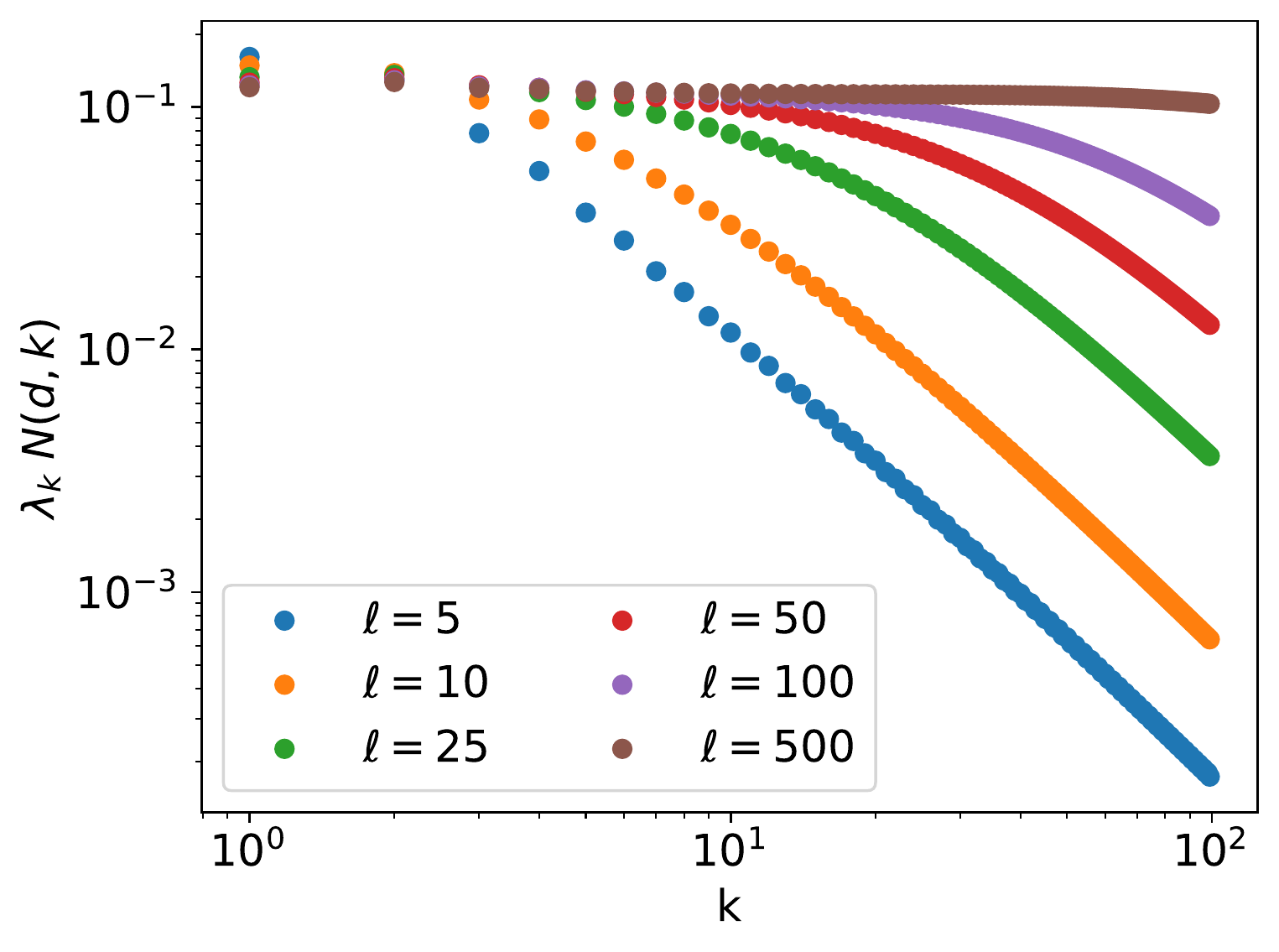}
    \caption{Spectrum of fully connected ReLU NTK without bias for varying depth $\ell$. As the depth increases, the spectrum whitens, causing derivatives of lower order to have infinite variance. As $\ell \to \infty$, $\lambda_k N(d,k) \sim 1$ implying that the kernel becomes non-analytic at the origin.}
    \label{fig:NTK_spectra}
\end{figure}

A plot of the RKHS spectra of fully connected ReLU NTK's of varying depth is shown in Figure \ref{fig:NTK_spectra}. As the depth increases, the spectrum becomes more white, eventually, the kernel's trace $\left< K(\mathbf{x},\mathbf{x})\right>_{\mathbf{x}} = \sum_k \lambda_k N(d,k)$ begins to diverge. Inference with such a kernel is equivalent to learning a function with infinite variance. Constraints on the variance of derivatives $\left< ||\nabla^n_{\mathbb{S}^{d-1}} f(\mathbf{x})||^2 \right>$ correspond to more restrictive constraints on the eigenspectrum of the RKHS. Specifically, $\lambda_k N(d,k) \sim \mathcal{O}( k^{-n-1/2} )$ implies that the $n$-th gradient has finite variance $\left< ||\nabla^n_{\mathbb{S}^{d-1}} f(\mathbf{x})||^2 \right> < \infty$.
\begin{proof}
By the representer theorem, let $f(\mathbf{x}) = \sum_{i=1}^p \alpha_i K(\mathbf{x},\mathbf{x}_i)$. 
By Green's theorem, the variance of the $n$-th derivative can be rewritten as
\begin{align}
\nonumber
   & \left< ||\nabla^n_{\mathbb{S}^{d-1}} f(\mathbf{x})||^2 \right> = \left< f(\mathbf{x}) (- \Delta_{\mathbb{S}^{d-1}})^n f(\mathbf{x})\right>
    \\
    \nonumber
    &\quad= \sum_{k k' m m' ij} \alpha_i \alpha_j \lambda_k \lambda_{k'} Y_{km}(\mathbf{x}_i) Y_{k'm'}(\mathbf{x}_j) 
    \\
    \nonumber
    &\qquad\times \left< Y_{km}(\mathbf{x}) (- \Delta_{\mathbb{S}^{d-1}})^n Y_{k'm'}(\mathbf{x}) \right>
    \\
    \nonumber
    &\quad= \sum_{k i j} \lambda_k^2 k^n(k+d-2)^n N(d,k) \alpha_i \alpha_j Q_{k}(\mathbf{x}_i^\top \mathbf{x}_j) 
    \\
    &\qquad\leq C p^2 (\alpha^*)^2 \sum_{k} \lambda_k^2 k^n(k+d-2)^n N(d,k)^2,
\end{align}
where $\alpha^* = \max_{j} |\alpha_j|$ and $|Q_{k}(z)| \leq C N(d,k)$ for a universal constant $C$. A sufficient condition for this sum to converge is that $\lambda_k^2 k^n(k+d-2)^n N(d,k)^2 \sim \mathcal{O}(k^{-1})$ which is equivalent to demanding $\lambda_k N(d,k) \sim \mathcal{O}(k^{-n-1/2})$ since $(k+d-2)^n \sim k^n$ as $k \to \infty$.
\end{proof}

\section{Decomposition of Risk for Numerical Experiments}

As we describe in Section \ref{sec:reg_expts} of the main text, the teacher functions for the kernel regression experiments are chosen as
\begin{equation}
    f^*(\mathbf{x}) = \sum_{i=1}^{p'} \overline{\alpha_{i}} K(\mathbf{x},\mathbf{\overline{x}}_i),
\end{equation}
where the coefficients $\overline{ \alpha_i} \sim \mathcal{B}(1/2)$ are randomly sampled from a centered Bernoulli distribution on $\{\pm 1\}$ and the points $\mathbf{\overline{x}}_i \sim p(\mathbf{x})$ are drawn from the same distribution as the training data. In general $p'$ is not the same as the number of samples $p$. Choosing a function of this form is very convenient for producing theoretical predictions of mode errors as we discuss below.

\subsection{Theoretical Mode Errors}

Since the matrix elements $\left< \mathbf{G}_{\rho \rho}^2 \right>$ are determined completely by the kernel eigenvalues $\{\lambda_\rho\}$, it suffices to calculate the diagonal elements of $\mathbf{D}$ to find the generalization error. For the teacher function sampled in the way described above, there is a convenient expression for $ \mathbf{D}_{\rho \rho}$.

The teacher function admits an expansion in the basis of kernel eigenfunctions
\begin{equation}
    f^*(\mathbf{x}) = \sum_\rho \overline{w}_\rho \psi_\rho(\mathbf{x}).
\end{equation}
Using the Mercer decomposition of the kernel we can identify the coefficients
\begin{equation}\label{kernel_teacher}
    f^*(\mathbf{x}) = \sum_{i=1}^{p'} \overline{\alpha}_{i} K(\mathbf{x},\mathbf{\overline{x}}_i) = \sum_{\rho} \Big( \sum_{i} \overline{\alpha_i} \psi_\rho(\mathbf{\overline{x}}_i) \Big) \psi_\rho(\mathbf{x})
\end{equation}
Comparing each term in these two expressions, we identify the coefficient of the $\rho$-th eigenfunction 
\begin{equation}
    \overline{w_\rho} = \sum_{i} \overline{\alpha_i} \psi_\rho(\mathbf{\overline{x}}_i).
\end{equation}

We now need to compute the $D_{\rho\rho}$, by averaging $\overline{w}_\rho^2$ over all possible teachers 
\begin{align}\label{d_expt}
\nonumber
    D_{\rho\rho} &= \frac{1}{\lambda_\rho} \left< \overline{w}_\rho^2 \right> = \frac{1}{\lambda_\rho} \sum_{ij} \left< \alpha_i \alpha_j \right> \left< \psi_\rho(\mathbf{\overline{x}}_i) \psi_\rho(\mathbf{\overline{x}}_j) \right>
    \\
    &= \frac{1}{\lambda_\rho} \sum_i \left<\psi_\rho(\mathbf{\overline{x}}_i) \psi_\rho(\mathbf{\overline{x}}_i) \right> = \frac{p'\lambda_\rho}{\lambda_\rho} = p',
\end{align}
since $\left< \psi_\rho(\mathbf{x}) \psi_\rho(\mathbf{x}) \right> = \lambda_\rho$. Thus it suffices to calculate $\frac{\partial}{\partial v} g_\rho(p,v)$ for each mode and then compute mode errors with
\begin{equation}
    E_\rho = - d_\rho \frac{\partial g_\rho(p,v)}{\partial v}|_{v=0},
\end{equation}
where $\frac{\partial g_\rho}{\partial v}|_{v=0}$ is evaluated in terms of the numerical solution for $t(p,0)$.

\subsection{Empirical Mode Errors}\label{SIRegressionModeError}

By the representer theorem, we may represent the student function as $f(\mathbf{x}) = \sum_{i=1}^P \alpha_i K(\mathbf{x},\mathbf{x}_i)$. Then, the generalization error is given by
\begin{align}
    \nonumber
    E_g &= \left< ( f(x) - f^*(x) )^2 \right> 
    \\
    \nonumber
    &= \sum_{\rho \gamma}   \lambda_\rho \lambda_\gamma \left( \sum_{j=1}^P \alpha_j \phi_\rho(x_j) - \sum_{i=1}^{P'} \overline{\alpha}_i \phi_\rho(\overline{x}_i)  \right) 
    \\
    \nonumber
    &\quad\left( \sum_{j=1}^P \alpha_j \phi_\gamma(x_j) - \sum_{i=1}^{P'} \overline{\alpha}_i \phi_\gamma(\overline{x}_i) \right) 
    \left< \phi_\rho(x) \phi_\gamma(x) \right>
    \\
    \nonumber
    &= \sum_\rho \lambda_\rho^2 \left( \sum_{j,j'} \alpha_{j} \alpha_{j'} \phi_\rho(x_j) \phi_\rho(x_j) \right.
    \\
    &\quad\left.- 2 \sum_{i,j} \alpha_j \overline{\alpha}_i \phi_\rho(x_j) \phi_\rho(\overline{x}_i)
     + \sum_{i,i'} \overline{\alpha}_i \overline{\alpha}_{i'} \phi(\overline{x}_i) \phi(\overline{x}_{i'}) \right).
\end{align}

On the $d$-sphere, by defining $E_k = \sum_{m=1}^{N(d,k)} E_{km}$ we arrive at the formula
\begin{align}
\nonumber
    E_k = \lambda_k^2 N(d,k) &\left( \bm{\alpha}^\top Q_k(\mathbf{X}^T \mathbf{X}) \bm{\alpha} 
    - 2 \bm{{\alpha}}^\top Q_k({\mathbf{X}}^T \mathbf{\overline{X}}) \bm{\overline{\alpha}}  \right.
    \\
    &\quad \left.+ \bm{\overline{\alpha}}^\top Q_k(\mathbf{\overline{X}}^T \mathbf{\overline{X}}) \bm{\overline{\alpha}}  \right).
\end{align}
We randomly sample the $\overline{\alpha}$ variables for the teacher and fit $\alpha = (\mathbf{K} + \lambda \mathbf{I})^{-1} \mathbf{y}$ to the training data. Once these coefficients are known, we can obtain empirical mode errors. 

\section{Neural Network Experiments}\label{SINeuralExperiment}

For the ``pure mode" experiments with neural networks, the target function was 
\begin{align}
\nonumber
     f^*(\mathbf{x}) &= \sum_{i=1}^{P'} \overline{\alpha}_i Q_k(\mathbf{x}^\top \mathbf{\overline{x}}_i) 
    \\
   & = \sum_{m=1}^{N(d,k)} \left( \sum_{i=1}^{P'} \overline{\alpha}_i  Y_{km}(\mathbf{\overline{x}}_i) \right) Y_{km}(\mathbf{x}),
\end{align}
whereas, for the composite experiment, the target function was a randomly sampled two layer neural network with ReLU activations 
\begin{align}
    f^*(\mathbf{x}) = \mathbf{\overline{r}}^\top \sigma( \mathbf{\overline{\Theta}} \mathbf{x}).
\end{align}
This target model is a special case of eq. \eqref{kernel_teacher} so the same technology can be used to compute the theoretical learning curves. We can use a similar trick as that shown in equation \eqref{d_expt} to determine $\overline{w}_\rho$ for the NN teacher experiment. Let the Gegenbauer polynomial expansion of $\sigma(z)$ be $\sigma(z) = \sum_{k=0}^\infty a_k N(d,k) Q_{k}(z)$.
Then the mode error for mode $k$ is $E_k = \frac{a_k^2}{\lambda_k^2} \left< g_k^2 \right>$ where $\left< g_k^2 \right>$ is computed with equation \eqref{mode_err}.

A sample of some training error and generalization errors from pure mode experiments are provided below in Figures \ref{fig:train_err} and \ref{fig:net_errs}.

\subsection{Hyperparameters}

The choice of the number of hidden units $N$ was based primarily on computational considerations. For two layer neural networks, the total number of parameters scales linearly with $N$, so to approach the overparameterized regime, we aimed to have $N \approx 10 p_{max}$ where $p_{max}$ is the largest sample size used in our experiment. For $p_{max}=500$, we chose $N=4000, 10000$. 

For the three and four layer networks, the number of parameters scales quadratically with $N$, making simulations with $N > 10^3$ computationally expensive. We chose $N$ to give comparable training time for the 2 layer case which corresponded to $N=500$ after experimenting with $\{100,250,500,1000,5000\}$. 

We found that the learning rate needed to be quite large for the training loss to be reduced by a factor of $\approx 10^{6}$. For the 2 layer networks, we tried learning rates $\{10^{-3},10^{-2},1,10,32\}$ and found that a learning rate of 32 gave the lowest training error. For the three and four layer networks, we found that lower learning rates worked better and used learning rates in the range from $[0.5,3]$.

\begin{figure*}[t]

\subfigure[3 Layer Training Loss; lr =2 ]{\label{fig:train_a1}\includegraphics[width=0.5\linewidth]{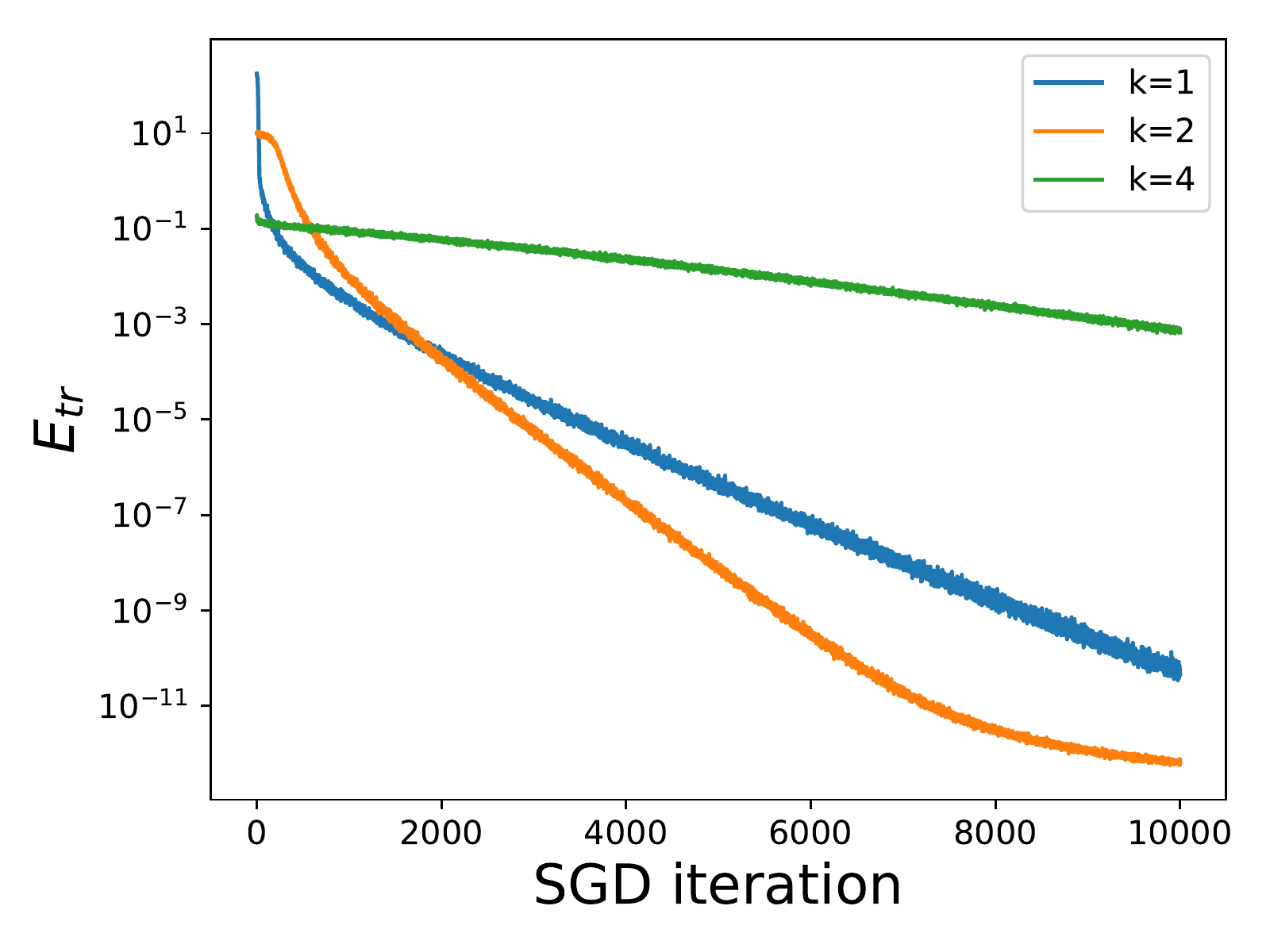}}
\subfigure[4 Layer Training Loss; lr = $0.5$]{\label{fig:train_a2}\includegraphics[width=0.5\linewidth]{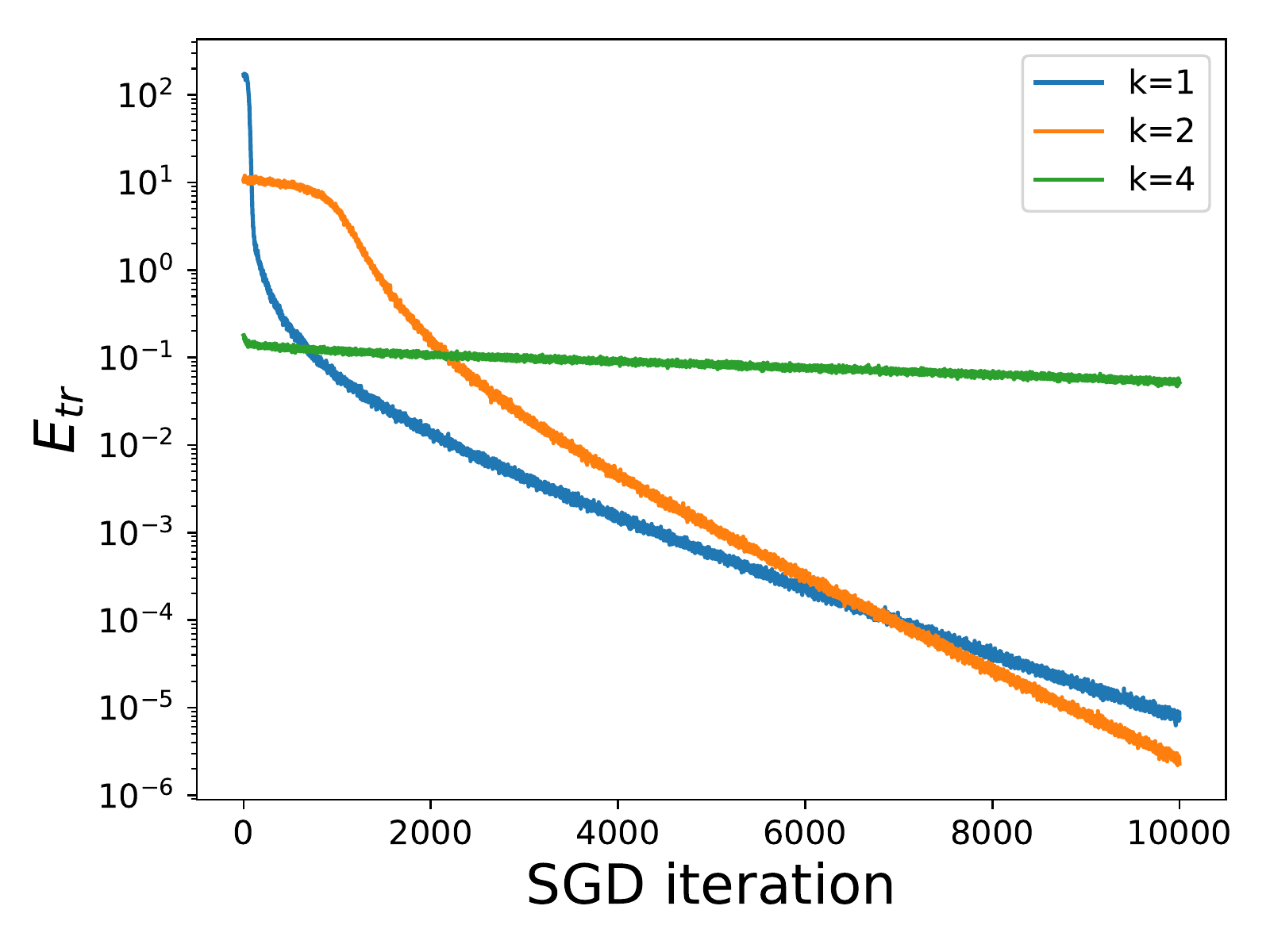}}
\vskip -8pt
\caption{Training error for different pure mode target functions on neural networks with 500 hidden units per hidden layer on a sample of size $p=500$. Generally, we find that the low frequency modes have an initial rapid reduction in the training error but the higher frequencies $k \geq 4$ are trained at a slower rate.}
\label{fig:train_err}
\end{figure*}

\begin{figure*}[t]
\subfigure[2 layer NN $N=4000$]{\label{sfig:net_a1}\includegraphics[width=0.33\linewidth]{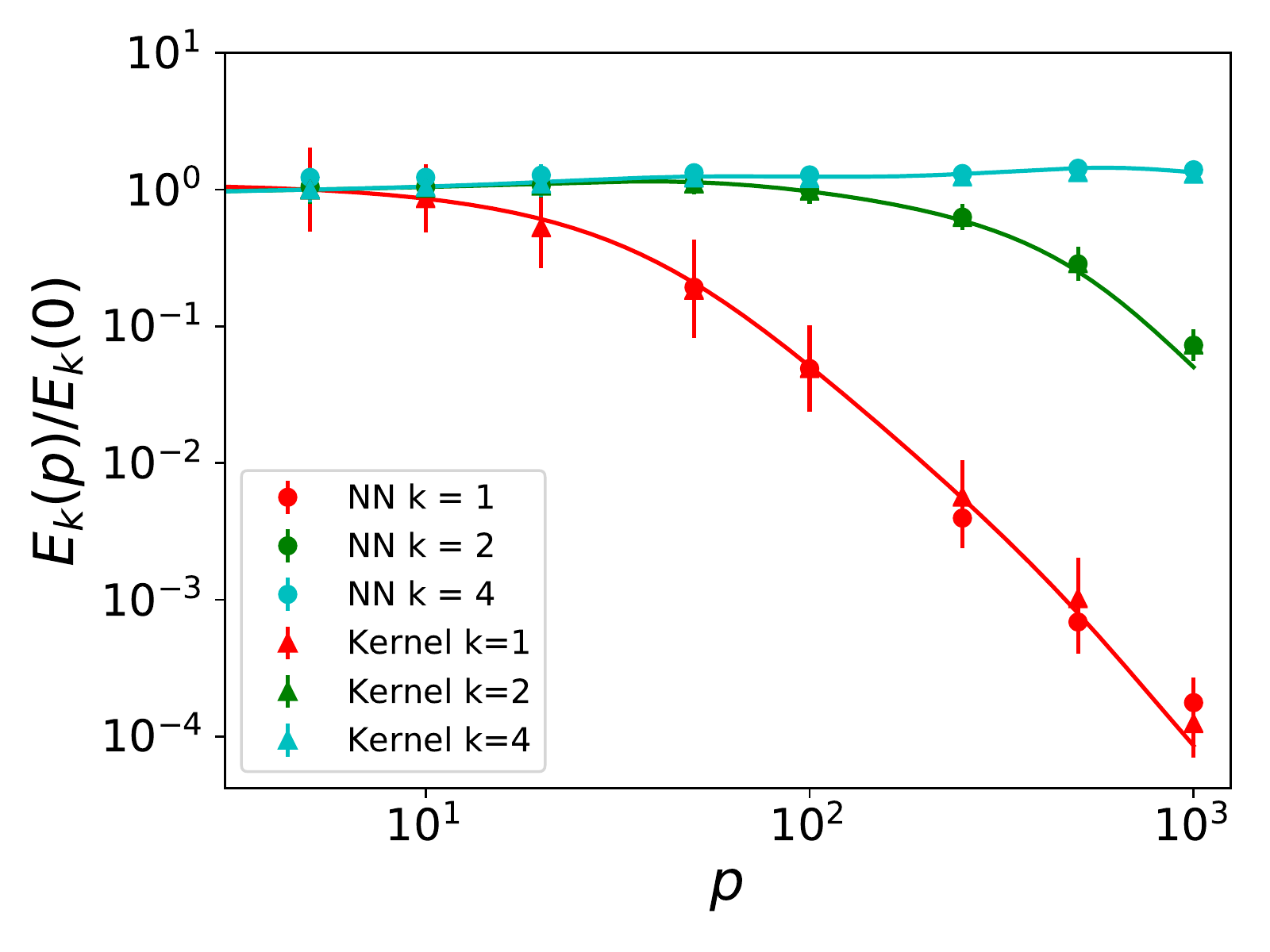}}
\subfigure[2 layer NN $N=10^4$]{\label{sfig:net_a2}\includegraphics[width=0.33\linewidth]{all_test_losses_rough_2_layer_d30_M10000_target10000_numiter7000_lr32.pdf}}
\subfigure[3 layer $N=500$ ]{\label{sfig:net_d1}\includegraphics[width=0.33\linewidth]{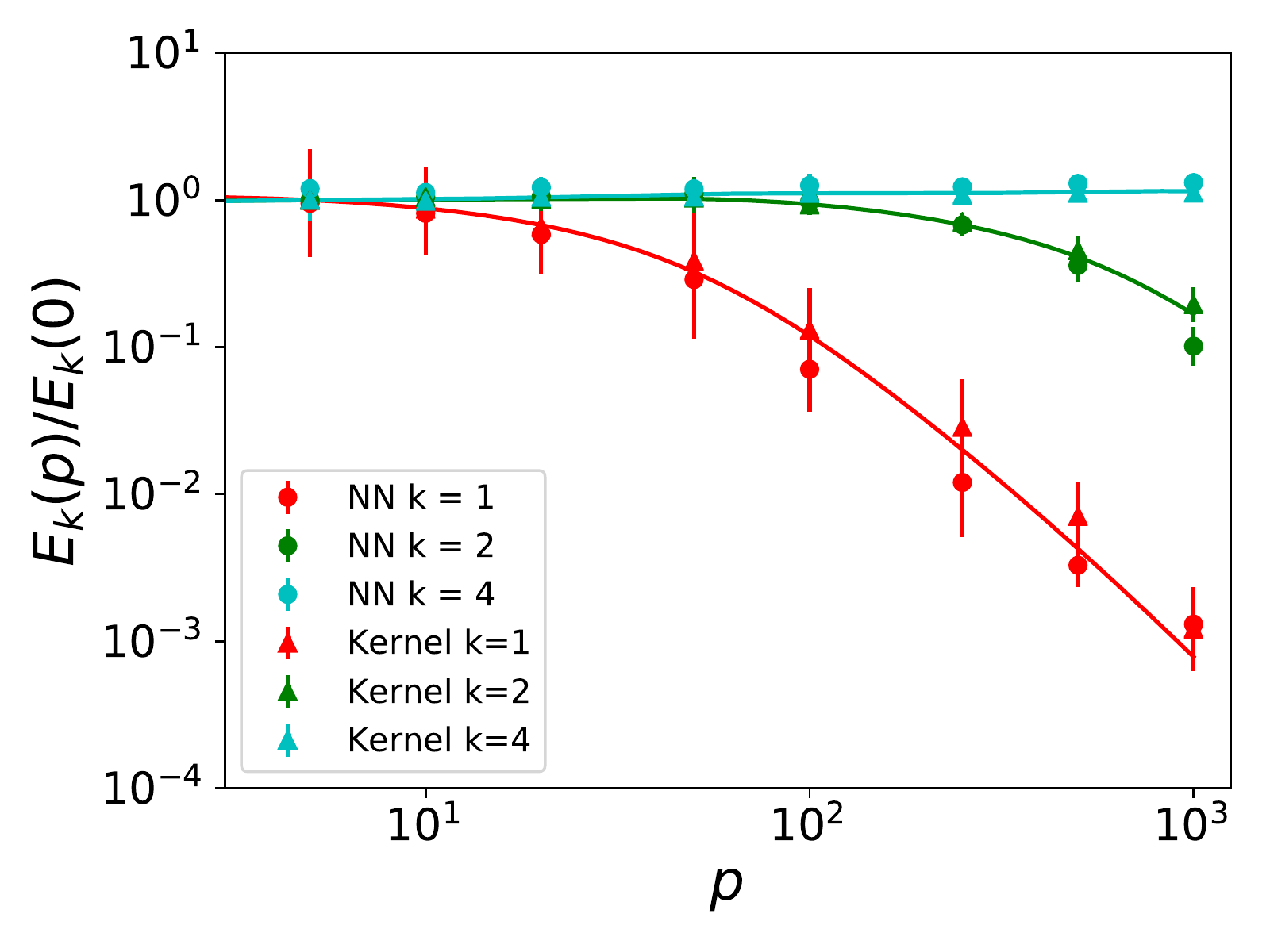}}
\subfigure[4 layer $N=500$ ]{\label{sfig:net_d2}\includegraphics[width=0.33\linewidth]{all_test_losses_rough_4_layer_d30_M500_target10000_numiter10000_lr0.pdf}}
\subfigure[2 Layer NN Student-Teacher; $N=2000$ ]{\label{sfig:net_c1}\includegraphics[width=0.33\linewidth]{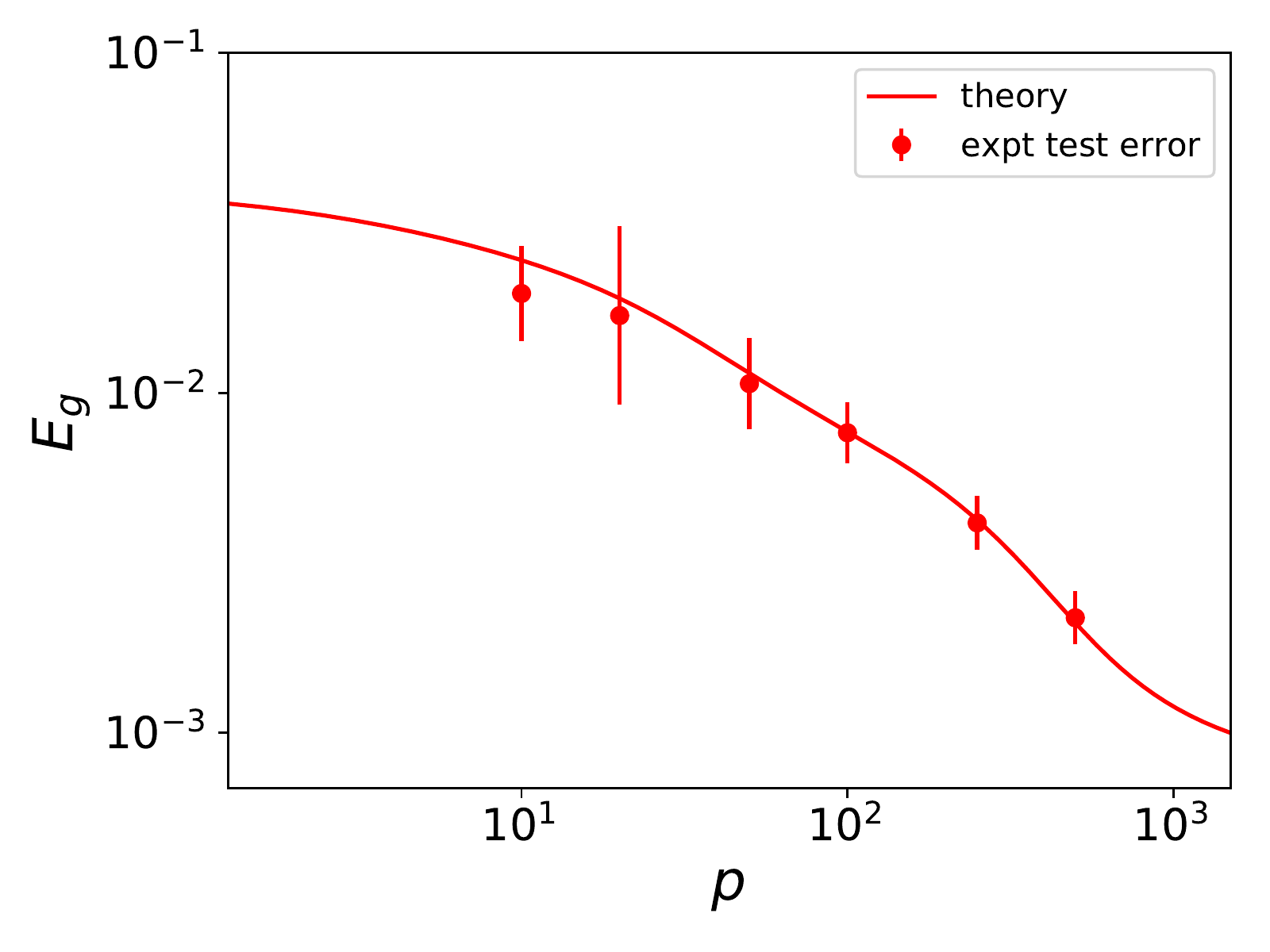}}
\subfigure[2 Layer NN Student-Teacher; $N=8000$ ]{\label{sfig:net_c2}\includegraphics[width=0.33\linewidth]{total_err_two_layer_NTK_M_8000_d_25.pdf}}

\vskip -8pt
 \caption{Learning curves for neural networks on ``pure modes" and on student teacher experiments. The theory curves shown as solid lines. For the pure mode experiments, the test error for the finite width neural networks and NTK are shown with dots and triangles respectively. Logarithms are evaluated with base 10.}
\label{fig:net_errs}
\end{figure*}

\section{Discrete Measure and Kernel PCA}\label{SI_KPCA}

We consider a special case of a discrete probability measure with equal mass on each point in a dataset of size $\tilde{p}$
\begin{equation}
    p(\mathbf{x}) = \frac{1}{\tilde{p}} \sum_{i=1}^{\tilde{p}} \delta(\mathbf{x}-\mathbf{x}_i).
\end{equation}
For this measure, the integral eigenvalue equation becomes
\begin{align}
    &\int d\mathbf{x}  \,p(\mathbf{x}) K(\mathbf{x},\mathbf{x}')  \phi_\rho(\mathbf{x})  \nonumber \\
    &\qquad=\frac{1}{\tilde{p}} \sum_{i=1}^{\tilde{p}} \int d\mathbf{x}\, \delta(\mathbf{x}-\mathbf{x}_i) K(\mathbf{x},\mathbf{x}') \phi_\rho(\mathbf{x})  \nonumber
    \\
    &\qquad = \frac{1}{\tilde{p}} \sum_{i=1}^{\tilde{p}} K(\mathbf{x}_i,\mathbf{x}') \phi_\rho(\mathbf{x}_i) = \lambda_\rho \phi_\rho(\mathbf{x}').
\end{align}
Evaluating $\mathbf{x}'$ at each of the points $\mathbf{x}_i$ in the dataset yields a matrix equation. Let $\mathbf{\Phi}_{\rho,i} = \phi_\rho(\mathbf{x}_i)$ and $\mathbf{\Lambda}_{\rho,\gamma} = \delta_{\rho,\gamma} \lambda_\rho$
\begin{equation}
    \mathbf{K} \mathbf{\Phi}^\top = \tilde{p}  \mathbf{\Phi}^\top \mathbf{\Lambda}.
\end{equation}

\end{document}